\documentclass[journal]{IEEEtran}

\usepackage{float}
\usepackage{times}
\usepackage{epsfig}
\usepackage{graphicx}
\usepackage{amsmath}
\usepackage{amssymb}
\usepackage{cite}
\usepackage{url}
\usepackage[noend]{algpseudocode}
\usepackage{array, bm, placeins, lipsum}
\usepackage{booktabs, color, enumerate, microtype, multirow, subcaption}
\usepackage[x11names,dvipsnames,table]{xcolor} 
\usepackage{arydshln}
\usepackage{verbatim}
\usepackage{enumitem}
\usepackage{pifont}
\usepackage{color}
\usepackage{makecell}
\newcolumntype{x}[1]{>{\centering\arraybackslash\hspace{0pt}}p{#1}}
\usepackage[table]{xcolor}

\newcommand{\PreserveBackslash}[1]{\let\temp=\\#1\let\\=\temp}
\newcolumntype{C}[1]{>{\PreserveBackslash\centering}p{#1}}
\newcolumntype{L}[1]{>{\PreserveBackslash\raggedright}p{#1}}

\makeatletter
\newif\if@restonecol
\makeatother
\usepackage[linesnumbered,ruled,vlined]{algorithm2e}
\usepackage{algpseudocode}

\usepackage{siunitx}
\ifCLASSINFOpdf
\else
\fi

\begin{document}
\title{Self-Supervised Masking for Unsupervised Anomaly Detection and Localization}

\author{Chaoqin~Huang, \textit{Student Member, IEEE}, Qinwei~Xu, \textit{Student Member, IEEE}, \\ Yanfeng~Wang, \textit{Member, IEEE}, Yu~Wang, \textit{Member, IEEE}, and Ya~Zhang, \textit{Member, IEEE}
\IEEEcompsocitemizethanks{\IEEEcompsocthanksitem 
This work is supported by the National Key Research and Development Program of China (No. 2020YFB1406801), 111 plan (No. BP0719010),  and STCSM (No. 18DZ2270700), and State Key Laboratory of UHD Video and Audio Production and Presentation. (Corresponding author: Yanfeng Wang and Ya Zhang)

C. Huang, Q. Xu, Y. Wang, Y. Wang, and Y. Zhang are with the Cooperative Medianet Innovation Center, Shanghai Jiao Tong University and Shanghai AI Laboratory, Shanghai 200240, China. (E-mail: \{huangchaoqin, qinweixu, wangyanfeng, yuwangsjtu, ya\_zhang\}@sjtu.edu.cn).}
}

\markboth{Journal of \LaTeX\ Class Files, July~2021}%
{Shell \MakeLowercase{\textit{\emph{et al.}\emph}}: Bare Demo of IEEEtran.cls for IEEE Journals}

\maketitle


\begin{figure*}
    \centering
    \includegraphics[width=1\textwidth]{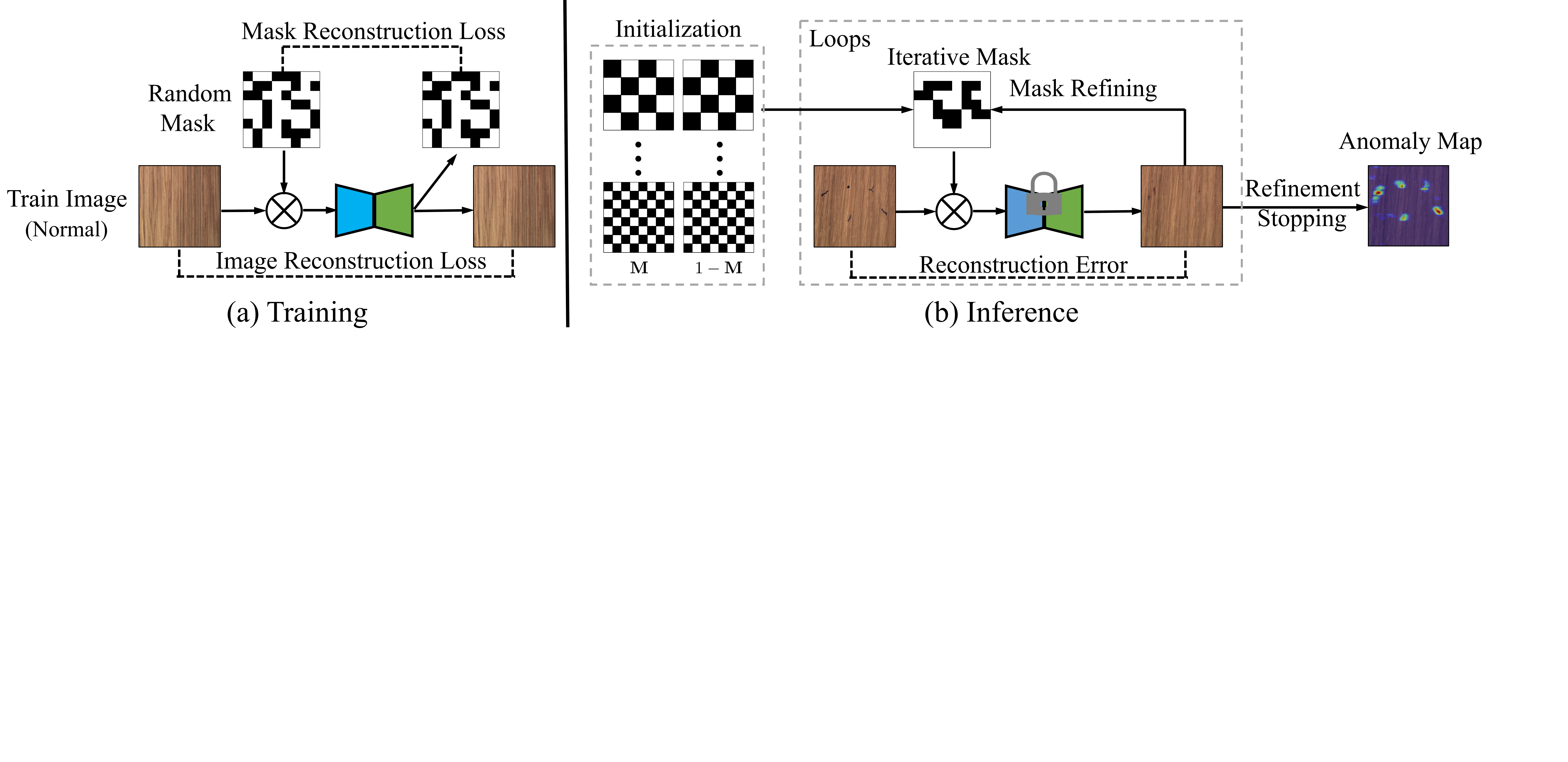}
    \caption{The overview of SSM. (a) During training, SSM leverages a conditional autoencoder to reconstruct the training images under randomly generated masks for only normal samples and also reconstruct the randomly generated masks, under a self-supervised learning paradigm. (b) During inference, SSM locates anomalies with a progressive mask refinement approach. An iterative mask is refined according to the feedback of the anomaly scores provided by the conditional autoencoder. SSM progressively uncovers the normal regions and finally locates the anomalous regions.}
    \label{fig:overview}
\end{figure*}

\begin{figure}
    \centering
    \includegraphics[width=0.49\textwidth]{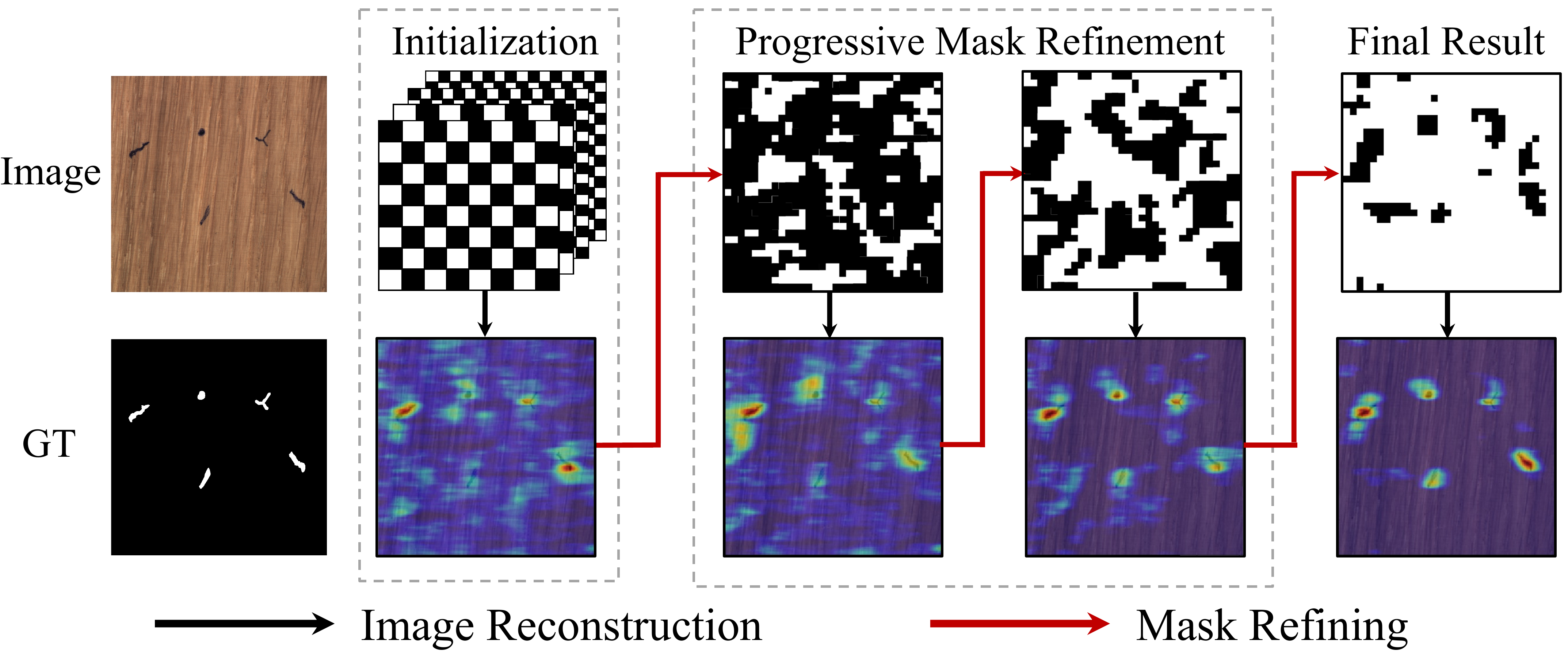}
    \caption{Results of anomaly localization during the inference stage with SSM on the MVTec AD dataset. During the process of progressive mask refinement, SSM continually narrows the scopes of both masks (top) and localization maps (bottom) to the anomalous regions.}
    \vspace{-8pt}
    \label{fig:introcase}
\end{figure}

\begin{abstract}
Recently, anomaly detection and localization in multimedia data have received significant attention among the machine learning community. In real-world applications such as medical diagnosis and industrial defect detection, anomalies only present in a fraction of the images. To extend the reconstruction-based anomaly detection architecture to the localized anomalies, we propose a self-supervised learning approach through \emph{random masking} and then \emph{restoring}, named \emph{\underline{S}elf-\underline{S}upervised \underline{M}asking} (SSM) for unsupervised anomaly detection and localization. SSM not only enhances the training of the inpainting network but also leads to great improvement in the efficiency of mask prediction at inference. Through random masking, each image is augmented into a diverse set of training triplets, thus enabling the autoencoder to learn to reconstruct with masks of various sizes and shapes during training. To improve the efficiency and effectiveness of anomaly detection and localization at inference, we propose a novel progressive mask refinement approach that progressively uncovers the normal regions and finally locates the anomalous regions. The proposed SSM method outperforms several state-of-the-arts for both anomaly detection and anomaly localization, achieving 98.3\% AUC on Retinal-OCT and 93.9\% AUC on MVTec AD, respectively.
\end{abstract}


\IEEEpeerreviewmaketitle

\section{Introduction}
\IEEEPARstart{A}{nomaly} detection and localization in multimedia data have received significant attention among the machine learning community, with broad application in medical diagnosis~\cite{baur2018deep,ouardini2019towards,tuluptceva2020anomaly,zhang2020viral}, defect detection in the factories~\cite{matsubara2018anomaly}, credit card fraud detection~\cite{phua2010comprehensive}, and autonomous driving~\cite{eykholt2018robust}.
For most of the above applications, anomalous samples are remarkably scarce in the population. It is often prohibitive to collect a representative set of anomalous samples. As a result, many studies \cite{Sabokrou2018Adversarially,perera2019ocgan,zhang2020p} have resorted to learning in the unsupervised setting, \emph{i.e.,} training with normal samples only.

Along this line, previous studies attempt to first model the normal distribution through either one-class classification-based approaches~\cite{scholkopf2001estimating,lee2017training,ruff2018deep}, reconstruction-based approaches~\cite{zong2018deep,deecke2018image,gong2019memorizing}, or self-supervision-based approaches~\cite{golan2018deep,wang2019effective,fye2020ARNet}, and then detect the anomalies by identifying samples with different distributions than the models. With the recent advances in deep neural networks, reconstruction-based approaches have received increasing attention and shown great promise for unsupervised anomaly detection. The network, trained with normal data only, is assumed not generalizable to abnormal samples, and thus leads to high reconstruction errors for anomalous samples. 

In real-world applications such as medical diagnosis~\cite{kermany2018identifying} and industrial defect detection~\cite{bergmann2019mvtec}, anomalies only present in a fraction of the images. To extend the reconstruction-based architecture to the localized anomalies, a few recent methods~\cite{yan2021learning,li2020superpixel} leverage image inpainting to locate anomalies from their surrounding context, with the assumption that the difference between the masked region and its corresponding restoration is significant for anomalous regions. Inpainting here needs to solve the dual tasks of \emph{masking} possible anomalies and \emph{restoring} the masked regions. While accurate restoration is critical to finding the masks for anomalies, it also depends on the precise masking of anomalies. To break the circular dependency between masking and restoring, SCADN~\cite{yan2021learning} introduces a fixed set of striped masks in multiple scales, and SMAI~\cite{li2020superpixel} leverages super-pixel segmentation to generate candidate masks, so that they can focus on fitting the restoration networks. Despite their promising results for anomaly localization, both methods require brutally traversing all possible masks, which is prohibitively time-consuming for real-world applications.

This paper explores a self-supervised learning approach through \emph{random masking} and then \emph{restoring}, named \emph{\underline{S}elf-\underline{S}upervised \underline{M}asking} (SSM) hereafter. The key difference between SSM and the above inpainting-based approaches~\cite{yan2021learning,li2020superpixel} mainly lies in the way of masking, which not only enhances the training of the inpainting network but also greatly improves the efficiency of mask prediction at test time. 
\textbf{(i)} During each training epoch, a random mask is generated on-the-fly for each image, and the masked image is then fed to a conditional autoencoder with two prediction heads, one for image reconstruction and the other for mask reconstruction (Figure~\ref{fig:overview} (a)). By use of random masking, each image is augmented into a diverse set of training triplets \emph{$<$masked image, mask, image$>$}, thus enabling the autoencoder to learn to reconstruct with masks of various sizes and shapes.
\textbf{(ii)} During the inference, as it is impossible to brutally traverse all possible masks (\emph{i.e.}, $2^n$ possible masks for an image of $n$ pixels),  a novel \emph{progressive mask refinement} approach is introduced to improve inference efficiency (Figure~\ref{fig:overview} (b)). SSM starts with a pair of complementary checkered masks, which jointly determine an initial mask according to the reconstruction results. The masks are iteratively refined and shrunk to the likely anomaly regions based on the reconstruction errors. Figure~\ref{fig:introcase} provides an illustration of the progressive anomaly localization process. To deal with anomalies of various shapes and sizes, the progressive mask refinement process is performed with initial masks of multiple scales and their ensemble results are used to detect the anomaly.

To validate the effectiveness of SSM, we experiment with two popular benchmark datasets, Retinal-OCT~\cite{kermany2018identifying} for medical diagnosis and MVTec AD~\cite{bergmann2019mvtec} for industrial defect localization. The experimental results have shown that SSM outperforms a number of state-of-the-art methods for both anomaly detection and localization, achieving 98.3\% anomaly detection AUC on Retinal-OCT~\cite{kermany2018identifying} and 93.9\% anomaly localization AUC on MVTec AD~\cite{bergmann2019mvtec}, respectively. 

The main contributions of the paper are summarized as follows:
\begin{itemize}
    \item We propose a novel masking \& restoring framework for unsupervised anomaly detection and localization named SSM. Through random masking, SSM enables the autoencoder to reconstruct the data with masks of various sizes and shapes, thus leading to a more powerful representation learning.
    \item To further improve the efficiency and effectiveness of anomaly detection and localization at inference, we propose a novel progressive mask refinement approach that progressively uncovers the normal regions and finally locates the anomalous regions.
\end{itemize}

\section{Related Works}
\subsection{Unsupervised Anomaly Detection}
Anomaly detection can be roughly divided into two classes: anomalous human behavior detection in videos~\cite{Kiran2018An, chu2018sparse, xu2018anomaly, xu2019video, sabokrou2016video, sabokrou2018avid, sabokrou2020deep, sabokrou2018deep, sabokrou2017deep, Sabokrou2018Adversarially} and anomaly detection in still images (or outlier data detection). In this paper, we focus on anomaly detection and localization in images, especially for medical diagnosis~\cite{kermany2018identifying} and industrial defect detection~\cite{bergmann2019mvtec}. Compared with the supervised approaches, unsupervised anomaly detection and localization is to train with normal samples only, without any anomalous data, and no image-level annotation or pixel-level annotation is provided. Since no auxiliary information for anomalies is provided, approaches like zero-shot object detection~\cite{ankan2018zero,yan2022semantics-guided} are also infeasible for unsupervised anomaly detection. Under the unsupervised setting, the majority of the research in image anomaly detection can be broadly categorized as one-class classification-based approaches, reconstruction-based approaches, generative adversarial network (GAN)-based approaches, and self-supervision-based approaches.

\subsubsection{One-class Classification-based Approaches} 
One-class classification is referred to as the problem of learning a description of a set of data instances to detect whether new instances conform to the training data or not. It assumes that all normal instances can be summarized by a compact model, to which anomalies do not conform~\cite{Eskin2000Anomaly, Yamanishi2000On, Rahmani2017Coherence, Xu2012Robust}. In OC-SVM~\cite{scholkopf2001estimating}, the normal samples are mapped to the high-dimensional feature space through a kernel function to get better aggregated. It learns a hyperplane that maximizes a margin between training data instances and the origin. To better aggregate the mapped data in latent space, Deep SVDD~\cite{ruff2018deep} optimizes the neural network by minimizing the volume of a hyper-sphere that encloses the network representations of the data. However, the one-class models may fail for datasets with complex distributions within the normal class. 
 
\subsubsection{Reconstruction-based Approaches}
Reconstruction-based anomaly detection approaches~\cite{an2015variational,xia2015learning,schlegl2017unsupervised,zong2018deep,deecke2018image} aim to learn the low-dimensional feature representation space on which the given normal data instances can be well reconstructed. The heuristic for using this technique in anomaly detection is that the learned feature representations are trained to learn regularities of the data. From this representation, anomalies are difficult to be reconstructed and thus have large reconstruction errors. DAE~\cite{Sakurada2014Anomaly} firstly applies the autoencoder to anomaly detection. To improve DAE, Nicolau \emph{et al.}~\cite{nicolau2016hybrid} introduce a density estimator, Kernel Density Estimation (KDE)~\cite{nicolau2016one}, to model the density from the hidden layer of autoencoders. By placing a threshold on the density of the normal data, query points below the threshold are classed as anomalies.

Recently, a deep autoencoder is adopted to improve the feature representation abilities~\cite{zhou2017anomaly,zavrtanik2021reconstruction,zhou2020encoding,zhou2021memorizing}. In the same vein, MemAE~\cite{gong2019memorizing} augments the autoencoder with a memory module to highlight reconstructed errors on anomalies. To capture the information of image texture and structure, P-Net~\cite{zhou2020encoding} proposes to leverage the relation between the image texture and structure to enlarge the structure difference, which can also be used as a metric for normality measurement. Based on~\cite{zhou2020encoding}, MemSTC~\cite{zhou2021memorizing} proposes a structure-texture correspondence memory module to reconstruct image texture from its structure, where a memory mechanism is used to characterize the mapping from the normal structure to its corresponding normal texture.

To detect anomalies that present in a fraction of an image, recent methods~\cite{yan2021learning,li2020superpixel} leverage image inpainting to extend the reconstruction-based approach to locate anomalies from their surrounding context. The difference between the masked region and its corresponding restoration is assumed to be significant for anomalous regions. SCADN~\cite{yan2021learning} leverages a fixed set of striped masks in multiple scales and traverses all possible masks at inference. Similarly, SMAI~\cite{li2020superpixel} leverages a superpixel masking and inpainting framework to identify and locate anomalies, where an inpainting module is trained to learn the spatial and texture information of the normal samples through random superpixel masking and restoration. Although the superpixel masks in SMAI may be more appropriate than the striped masks in SCADN, it takes longer for inference due to the need of traversing more possible masks (77 possible masks in total). This paper introduces a progressive mask refinement approach to avoid brutally traversing all possible masks, which improves inference efficiency.

\subsubsection{Generative Adversarial Network-based Approaches} 
By assuming that normal data instances can be better generated than anomalies from the latent feature space of the generative network, adversarial training is employed~\cite{zenati2018efficient,Sabokrou2018Adversarially,sabokrou2020deep}. These approaches generally aim to learn a latent feature space of a generative network so that the latent space well captures the normality underlying the given data. Along this line, to improve the generator's robustness against noises, ALOCC adds Gaussian noises to the inputs to form the training normal samples~\cite{Sabokrou2018Adversarially}. GANomaly~\cite{akccay2019skip} leverages another encoder to embed the generated results to a subspace, and calculates the anomaly scores in the subspace but not in the image space like ALOCC. OCGAN~\cite{perera2019ocgan} further applies two adversarial discriminators and a classifier on a denoising autoencoder. By adding constraints and forcing latent codes to reconstruct examples like the normal data, anomalies show higher reconstruction errors. Similarly, adVAE~\cite{wang2020advae} employs adversarial training within a variational autoencoder framework under the assumption that normal and anomalous data follows different Gaussian distributions. However, the generator networks can be misled and thus generate data instances out of the manifold of normal instances, especially when the distribution of the given dataset is complex or the training data contains unexpected outliers.

\subsubsection{Self-supervision-based Approaches}
Recently, self-supervised learning has been widely used as it benefits many downstream tasks like classification~\cite{noroozi2016unsupervised,gidaris2018unsupervised}, detection~\cite{doersch2015unsupervised,yan2022semantics-guided}, segmentation~\cite{gidaris2018unsupervised}, and tracking~\cite{yuan2021self-supervised}. Among various self-supervised tasks, image restoration or inpainting has been widely adopted, where the network is forced to learn rich and robust feature embeddings. Pathak \emph{et al.}~\cite{pathak2016context} propose to remove arbitrary shapes from the input images. Those shapes are obtained as objects in the PASCAL VOC 2012 dataset~\cite{Pascal} and pasted in arbitrary places in the other images. In~\cite{denton2016semi}, a low resolution but intact version of the original image is further fed to the network to guide the restoration. Based on~\cite{pathak2016context}, \cite{jenni2018self} further proposes to remove and restore the internal representations and designs a feature attention module to improve the feature robustness.

Self-supervision-based anomaly detection approaches~\cite{golan2018deep,fye2020ARNet,bergmann2020uninformed} learn the representations of the normal data under a self-supervised learning paradigm with different self-supervisions~\cite{doersch2015unsupervised,noroozi2016unsupervised,gidaris2018unsupervised}. Models are optimized with different surrogate tasks. Then anomalies can be separated under the assumption that anomalies will result differently in the corresponding surrogate task. Similarly, GeoTrans applies dozens of image geometric transforms and creates a self-labeled dataset for transformation classification~\cite{golan2018deep,gidaris2018unsupervised}. It assumes that transformations applied on anomalous data can not be classified properly. Wang \emph{et al.}~\cite{wang2019effective} apply Jigsaw puzzles~\cite{noroozi2016unsupervised} to extend the above self-labeled dataset. \cite{CutPaste} learns representations by classifying data between different types of CutPaste, a set of data augmentations, and then utilizes a sliding window for anomaly localization although it is time-consuming. ARNet~\cite{fye2020ARNet} applies image restoration as the self-supervision, assuming that the model is able to learn semantic features during the restoration process. GP~\cite{GP} proposes a patch-based approach that considers both the global and local information and utilizes the discrepancy between global and local features as the anomaly score. Based on the knowledge distillation approach, US~\cite{bergmann2020uninformed} and MKD~\cite{MKD} train the student networks to regress the output of a descriptive teacher network that was pre-trained on a large dataset. Anomalies are detected when the outputs of the student networks differ from that of the teacher network. This happens when they fail to generalize outside the manifold of anomaly-free training data. In this paper, self-supervised learning is applied in the proposed approach for both conditional image reconstruction and mask reconstruction.

\begin{figure*}[t]
    \centering
    \includegraphics[width=1.0\textwidth]{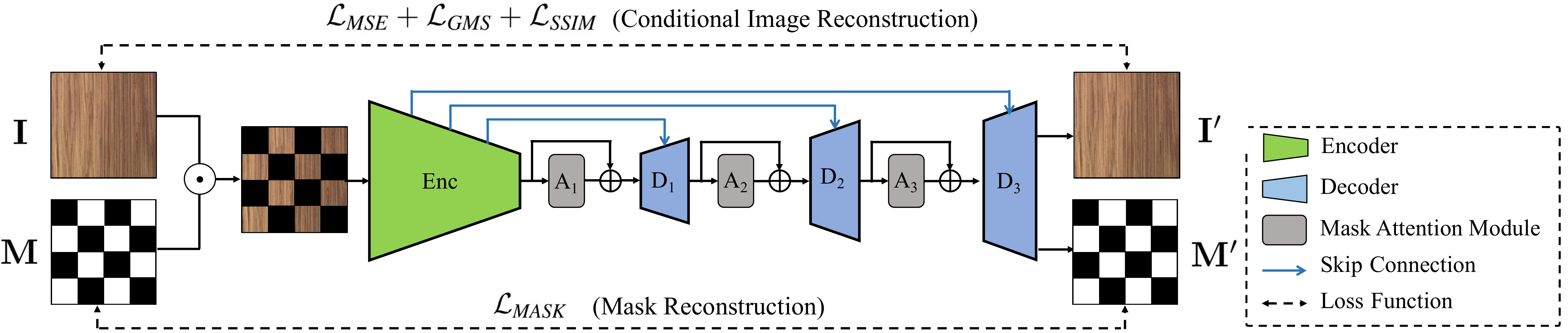}
    \caption{Model architecture of the conditional autoencoder for the self-supervised masking training. An autoencoder $\{\text{Enc},\text{D}_1,\text{D}_2,\text{D}_3\}$ takes $\mathbf{I}\odot \mathbf{M}$ as input and outputs both reconstructed images $\mathbf{I}'$ (top) and the reconstructed masks $\mathbf{M}'$ (bottom). Some mask attention modules $\{\text{A}_1 , \text{A}_2 , \text{A}_3\}$ distributed across the layers of the decoder (see Figure~\ref{fig:mam} for more details).}
    \label{fig:archi}
\end{figure*}

\begin{figure}[t]
    \centering
    \includegraphics[width=0.48\textwidth]{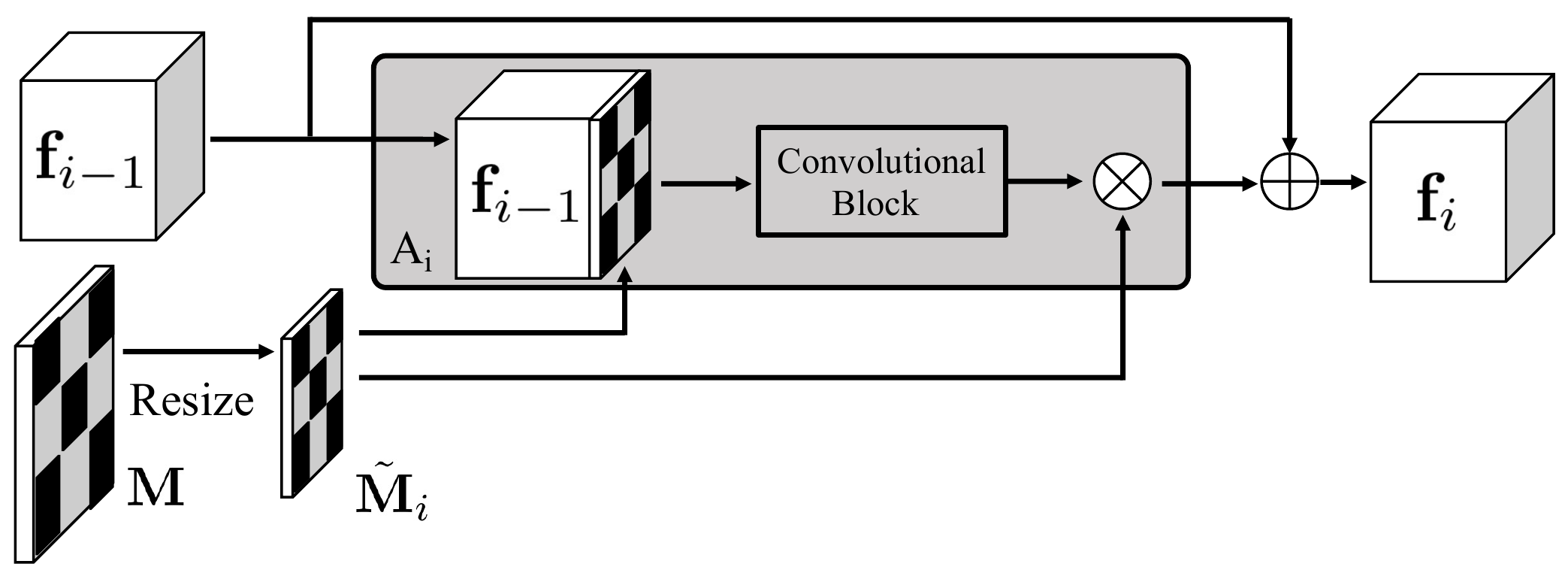}
    \caption{Mask attention module (MAM) of conditional autoencoder for self-supervised masking training. The mask attention module $\text{A}_i$ takes the feature $\mathbf{f}_{i-1}$ as input and output $\mathbf{f}_{i}$ for the next decoder part. Following~\cite{jenni2018self}, we use a residual block design and gate the output of the last convolutional layer with the resized mask $\tilde{\mathbf{M}}_i$.}
    \label{fig:mam}
\end{figure}

\subsection{Progressive Refinement}
The progressive refinement network has been explored in many supervised tasks, such as supervised image matting~\cite{MGM}, person re-identification~\cite{zhang2021seeing}, and temporal action detection~\cite{liu2020progressive}, motivated by the thought of progressive learning~\cite{huang2019tpckt,fayek2020progressive}.
For example, PBRNet~\cite{liu2020progressive} is equipped with three cascaded detection modules for progressive localizing action boundaries more and more precisely. MGMatting~\cite{MGM} proposes a progressive refinement network for image matting, which encourages the matting model to provide self-guidance to progressively refine the uncertain regions through the decoding process in multiple layers of the feature hierarchy. In this paper, we leverage progressive mask refinement for unsupervised anomaly detection. Different from the multiple feature hierarchy refinement for supervised tasks, the proposed progressive mask refinement is a procedure at inference, which takes the continuously refined mask as the input and reuses the trained conditional autoencoder for image reconstruction.

\section{Method}
The key novelties of SSM include \emph{random masking and restoring} at training and \emph{progressive mask refinement} at inference. As the training data are all normal, a random mask is generated to formulate the inpainting task for each image. A conditional autoencoder is then leveraged to restore the masked region. At inference, to avoid brutally traversing all possible masks, a progressive mask refinement approach is proposed to improve the efficiency of anomaly detection and localization.

\subsection{Random Masking} \label{sec:mask}
To define the inpainting area of the training images, each input image is decomposed into $\frac{H}{k} \times \frac{W}{k}$ grids, where $H$ and $W$ are the height and width of the images, and $k$ here controls the granularity of the grid. Each grid consists of a square of $k \times k$ pixels and is set as the basic unit of the masks. The match of the grid size $k$ and the size of the anomaly is expected to significantly influence the performance of anomaly detection algorithms. If $k$ is much larger or much smaller than the anomaly, it is generally infeasible to obtain accurate reconstruction. Since anomalies could come in various sizes, and there is no way to know the size of the anomalies as prior, we here consider detecting with multiple scales by varying the values of $k$. Particularly, the size $k$ is sampled from a set $K = \{k_i\}_{i=1:N_k}$, where $N_k$ is the set cardinality. In our implementation, we use $K = \{4, 8, 16, 32\}$ as it covers a wide range of anomaly scales. 

To enlarge the exploration space of the masks, during each training epoch, a random mask is generated on-the-fly for each image. Each grid is then randomly chosen to be masked or to be kept, and the resulting mask matrix is denoted as $\mathbf{M}$. In this way, a diverse set of random masks with various sizes and shapes are generated. Through such random masking, each image is augmented into a diverse set of training triplets $<\tilde{\mathbf{I}}$, $\mathbf{M}$, $\mathbf{I}>$, where $\mathbf{I}$ is the input image, $\mathbf{M}$ is the generated spatial mask, $\tilde{\mathbf{I}} = \mathbf{I} \odot \mathbf{M}$ is the resulting masked image, and $\odot$ is the element-wise product in the spatial domain (the mask is replicated along the channel dimension), thus enabling to learn to reconstruct with masks of various sizes and shapes.

\subsection{Restoration Network}\label{sec:train}
The overall architecture of the restoration network is shown in Figure~\ref{fig:archi}. The backbone of the restoration network is a conditional autoencoder. Different from the vanilla autoencoder, the conditional autoencoder is used to encode unmasked regions and fill in masked regions under a certain mask matrix~\cite{yu2019free}. The condition lies in that the proposed image reconstruction network is mask-guided but not simply reconstructing the whole image. Different from vanilla image reconstruction-based anomaly detection methods, we assume that the discrepancy between the masked region and its corresponding restoration is significant for detecting anomalies.

For better representation learning, we further add a mask reconstruction branch in our network so that two prediction heads are associated with the conditional autoencoder, one for image reconstruction and the other for mask reconstruction:
\begin{equation}
        \mathbf{I', M'} = \text{Dec}[\text{Enc}(\mathbf{I}\odot \mathbf{M})],
\end{equation}
where $\text{Enc}$ is the encoder, $\text{Dec}$ is the decoder, $\mathbf{M}'$ is the reconstructed mask and $\mathbf{I}'$ is the reconstructed image. Skip-connections between the encoder and decoder are added to facilitate the backpropagation of gradients and improve the performance of image reconstruction. 

Inspired by~\cite{jenni2018self}, to improve the robustness and reconstruction ability of the model in the manner of self-supervised learning, we further add a mask attention module (MAM) to the conditional autoencoder. The architecture of MAM is shown in Figure~\ref{fig:mam}. The decoder $\text{Dec}$ is split into three sub-networks $\{\text{D}_1, \text{D}_2, \text{D}_3\}$, and the mask attention module, $\text{A} = \{\text{A}_1, \text{A}_2, \text{A}_3\}$, is added in front of each of the sub-networks for the decoder network, as shown in Figure~\ref{fig:archi}. For the $i$-th mask attention module $\text{A}_i$, we down-sample the mask $\mathbf{M}$ to $\tilde{\mathbf{M}}_i$ with the nearest neighbor method and match the spatial dimension of the corresponding input feature map $\mathbf{f}_{i-1}$. The output $\mathbf{f}_i$ of the $i$-th mask attention module $\text{A}_i$ is:
\begin{equation}
    \mathbf{f}_i = \mathbf{f}_{i-1} + \phi(\text{C}(\mathbf{f}_{i-1},\tilde{\mathbf{M}}_i)) \odot \tilde{\mathbf{M}}_i,
\end{equation}
where $\text{C}(\cdot,\cdot)$ is a concatenation layer and $\phi(\cdot)$ is a convolutional block following~\cite{jenni2018self}, and $\odot$ denotes the element-wise product in the spatial domain (the mask is replicated along the channels). With the mask attention module, the model pays more attention to learn the context feature, thus the model's image restoration ability can be significantly improved.

\subsection{Loss Functions}\label{sec:loss}
Given the input image $\mathbf{I}$, and the reconstructed image $\mathbf{I}'$, as the network only targets to restore the masked regions, the unmasked regions are copied from the original images with an identity function:
\begin{equation}
        \mathbf{\hat{\mathbf{I}}} = \mathbf{I}' \odot (1-\mathbf{M}) + \mathbf{I} \odot \mathbf{M}.
\end{equation}
For image reconstruction, we aim to measure the difference between $\mathbf{I}$ and $\hat{\mathbf{I}}$ and consider the following set of loss functions. 

\noindent \textbf{Mean Square Error:} The mean square error (MSE) is typically used for training an autoencoder. 
\begin{equation}
    \mathcal{L}_{\textit{MSE}} = \|\mathbf{I}-\hat{\mathbf{I}}\|^2_2.
\end{equation}
However, this loss assumes the independence between neighboring pixels, which may be incorrect in some situations. To solve this problem, this paper further introduces several losses that penalize structural differences between the reconstructed regions and the input regions, \emph{i.e.,} a gradient magnitude similarity (GMS) loss~\cite{xue2013gradient} and a structured similarity index (SSIM) loss~\cite{wang2004image}. SSIM and GMS are both patch similarity metrics that focus on different image properties. 

\noindent \textbf{Gradient Magnitude Similarity Loss:} The gradient difference map is defined as:
\begin{equation}
    \mathcal{L}_\textit{GMS} = \frac{1}{H\times W} \sum_{i=1}^{H} \sum_{j=1}^{W} \mathbf{1} - \textit{GMS}(\mathbf{I},\hat{\mathbf{I}})_{(i,j)},
\end{equation}
\begin{equation}
    \textit{GMS}(\mathbf{I},\hat{\mathbf{I}}) = \frac{1}{3}\sum_{c=1}^3 \textit{GMS}_c(\mathbf{I}^c,\hat{\mathbf{I}}^c)\in \mathbb{R}^{H\times W},
\end{equation}
where $\mathbf{1}$ is a matrix of ones. $\mathbf{I}^c$ and $\hat{\mathbf{I}}^c$ are $c$-th color channel of the original image and the reconstructed image, respectively. $\textit{GMS}_c(\mathbf{I}^c,\hat{\mathbf{I}}^c)\in \mathbb{R}^{H\times W}$ is the gradient magnitude similarity map for color channel $c$:
\begin{equation}
    \textit{GMS}_c(\mathbf{I}^c,\hat{\mathbf{I}}^c)=\frac{2 g(\mathbf{I}^c) g(\hat{\mathbf{I}}^c)+a}{g(\mathbf{I}^c)^{2}+g(\hat{\mathbf{I}}^c)^{2}+a},
\end{equation}
\begin{equation}
   g(\mathbf{I}^c) = \sqrt{(\mathbf{I}^c * \mathbf{h}_{x})^{2}+(\mathbf{I}^c * \mathbf{h}_{y})^{2}},
\end{equation}
where $a$ is a constant ensuring numerical stability. $\mathbf{h}_{x}$ and $\mathbf{h}_{y}$ are $3\times 3$ Prewitt filters along the $x$ and $y$ dimensions and $\ast$ is the convolution operation. The GMS loss is differentiable and has been widely used by image inpainting and image super-resolution tasks~\cite{zhu2020gan,shahsavari2021proposing}.

\noindent \textbf{Structured Similarity Index (SSIM):} The SSIM loss is defined as:
\begin{equation}
    \mathcal{L}_{\textit{SSIM}}(\mathbf{I}, \hat{\mathbf{I}})=
    \frac{1}{H\times W} \sum_{i=1}^{H} \sum_{j=1}^{W}  \mathbf{1}-\textit{SSIM}(\mathbf{I}, \hat{\mathbf{I}})_{(i, j)},
\end{equation}
where $\textit{SSIM}(\mathbf{I},\hat{\mathbf{I}})_{(i,j)}$ is the SSIM~\cite{wang2004image} value between two patches of $\mathbf{I}$ and $\hat{\mathbf{I}}$ centered at $(i, j)$.

The mask reconstruction simply adopts the $L_2$ loss:
\begin{equation}
    \mathcal{L}_{\textit{MASK}} = \|\mathbf{M}-\mathbf{M}'\|^2_2.
\end{equation}

Finally, the total loss for training SSM is formulated as:
\begin{equation}\label{eq:final}
    \mathcal{L} = \lambda_1\mathcal{L}_{\textit{MSE}} + \lambda_2\mathcal{L}_{\textit{GMS}}+\lambda_3\mathcal{L}_{\textit{SSIM}}+\lambda_4\mathcal{L}_{\textit{MASK}},
\end{equation}
where $\lambda_1$, $\lambda_2$, $\lambda_3$, and $\lambda_4$ are the individual loss weights.

\subsection{Progressive Mask Refinement at Inference}\label{sec:test}

At inference time, one additional challenge emerges, \emph{i.e.,} how to set up the mask for conditional reconstruction? It is desired that the mask covers only the anomalous regions but not the normal region, which is in fact a dilemma: if we have a perfect mask, we can correctly locate the anomalies; otherwise, the reconstruction result may not correctly manifest the anomalies. To break the dilemma, existing inpainting-based approaches~\cite{yan2021learning,li2020superpixel} simply brutally traverse all possible masks, but at the cost of limiting the search space for masks. In our case, with random masking, it is impossible to brutally traverse all possible masks. A novel progressive mask refinement approach, consisting of two stages, \emph{i.e.}, mask initialization and mask refinement, is introduced to improve the efficiency for inference.

Given the input image $\mathbf{I}$ and the reconstructed image $\hat{\mathbf{I}}$, an error function $f(\cdot,\cdot)$ is introduced
as follows.
\begin{equation} 
    f(\mathbf{I},\hat{\mathbf{I}}) = L_2(\mathbf{I},\hat{\mathbf{I}})+ (\mathbf{1}-\textit{GMS}(\mathbf{I},\hat{\mathbf{I}}))+ (\mathbf{1}-\textit{SSIM}(\mathbf{I},\hat{\mathbf{I}})),
    \label{eq:fscore}
\end{equation}
where $L_2(\cdot,\cdot)$, $\textit{GMS}(\cdot,\cdot)$, $\textit{SSIM}(\cdot,\cdot)$ are the per-pixel $L_2$, GMS and SSIM score maps, respectively. This error function is then leveraged for both anomaly score calculation and mask refinement. By calculating the per-pixel-based error scores and treating regions with large scores as potential anomalous, the masks are iteratively refined and gradually shrunk to the possible anomaly regions.

\begin{figure}[t]
    \centering
    \includegraphics[width=0.48\textwidth]{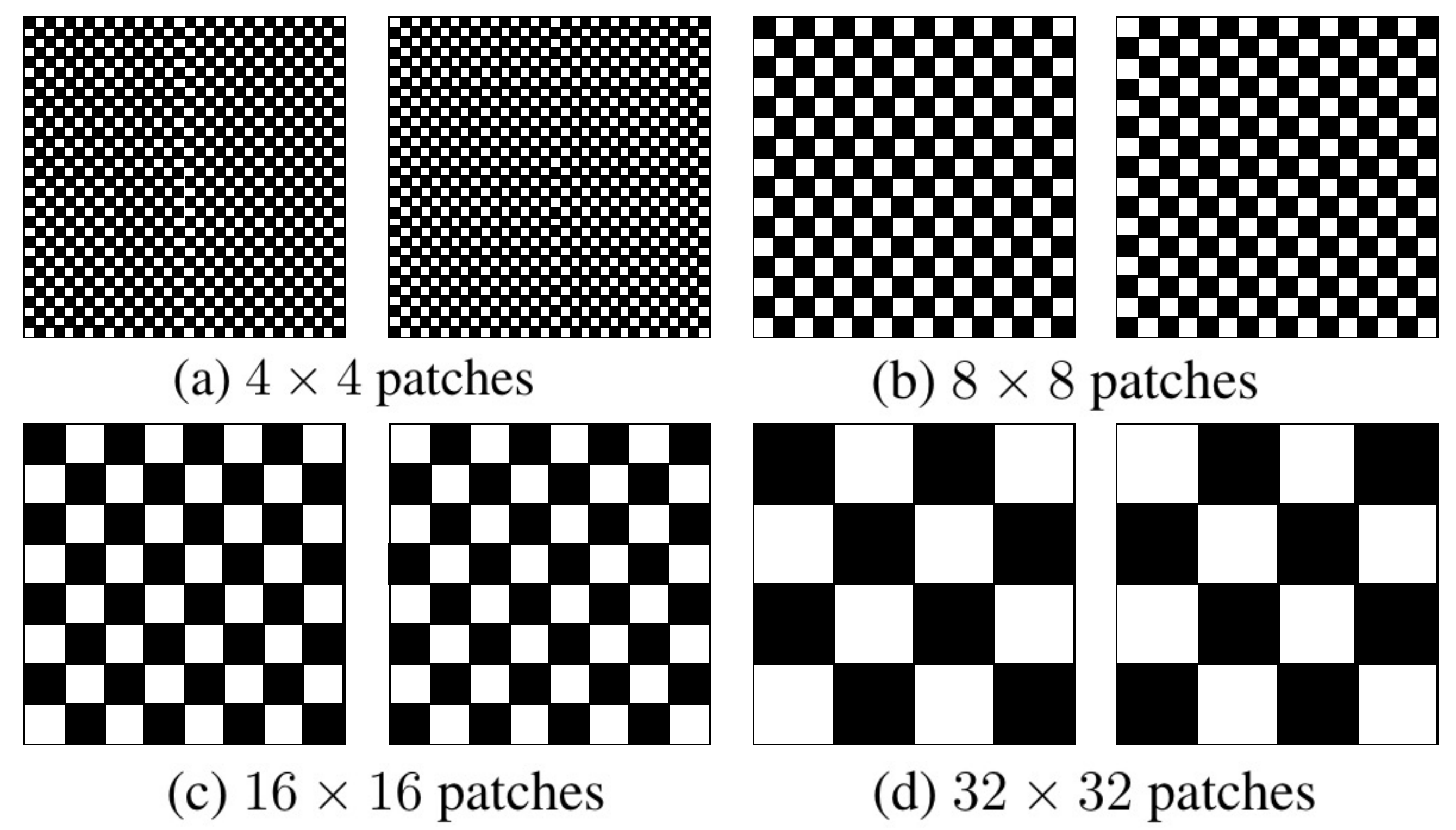}
    \caption{Mask initialization during the inference. These initialized masks are complementary with each other for each size of patches and thus cover all pixels in the image, which avoids missing possible anomalous areas.}
    \label{fig:maskinit}
\end{figure}

\subsubsection{Mask Initialization}
Given a test image, assuming no prior information about the anomalous region at the beginning, as initialization, we start with a set of multi-scaled masks $\mathcal{M}^{0}$, which consists of eight checkerboard-like matrices in different scales. Figure~\ref{fig:maskinit} shows the set of initialization masks used in our experiments, where the grid size $k\in K, K = \{4, 8, 16, 32\}$ in our experiments. For each grid size $k$, the initialization masks contain a pair of complementary masks which jointly cover all pixels in the image, thus avoiding missing any possible anomalous areas.

For each initialization mask, the model reconstructs the pixels in masked regions of the test images based on the corresponding pixels in un-masked regions. The anomaly score map is obtained for each reconstructed image using the error function $f(\cdot,\cdot)$ as defined in Eq. \ref{eq:fscore}. The score maps for different masks are average into a single one to obtain the initialized anomaly score map $\mathbf{S}^0$.

\subsubsection{Mask Refinement}
The purpose of mask refinement is to remove the masked areas likely corresponding to the normal regions, so that the conditional image reconstruction network pays more attention to the remaining anomalous regions. At each iteration, the reconstruction error map is leveraged to refine the mask, by considering regions with smaller errors as normal and removing them from the mask for the next iteration. 

When most of the regions covered by the mask are anomalous regions, providing more image information can not reduce the reconstruction error of anomalous regions significantly, and the corresponding mask is unchanged/converged. At this point, we terminate the inference stage, and obtain the final masks. Finally, when this approach ends, the mask is expected to cover only the anomalous parts of the image.

\begin{algorithm}[t]
  \caption{Progressive Mask Refinement}
  \label{alg:update}  
  \KwIn{input $\mathbf{I}\in \mathbb{R}^{H\times W\times C}$,\\ ~~~~~~~~~conditional autoencoder $\text{Dec}[\text{Enc}(\cdot)]$, \\
  ~~~~~~~~~anomaly score map calculation function $f(\cdot,\cdot)$,\\
  ~~~~~~~~~an initialized score map $\mathbf{S}^{0}$,\\
  ~~~~~~~~~patch size $k$, a threshold $\eta$.}
  \KwOut{anomaly score $\epsilon_k$ (for detection),\\
        ~~~~~~~~~~~anomaly score map $\mathbf{S}^k$ (for localization).}
    $\mathbf{S}^k = \mathbf{S}^{0}$\\ 
    $N_k = \frac{H}{k} \times \frac{W}{k}$ \hfill\# numbers of the patches\\
    split image $\mathbf{I}$ into $N_k$ $k\times k$ patches $p_1,p_2,\cdots,p_{N_k}$\\
    \Repeat{$\mathbf{M}$ converges}
    {
        \# mask refinement\\
        \For{each patch p}
        {
        $\epsilon_{p} = \frac{1}{k^2}\sum_{(x,y)\in p}\mathbf{S}^k_{(x,y)}$\\
        \For{each pixel $(x,y)$ in $p$}
            {
            $\mathbf{M}_{(x,y)}=\left\{\begin{matrix}0 & \text{ if } \epsilon_{p} > \eta\\ 1 & otherwise \end{matrix}\right.$\\
            }
        }
        \# score calculation\\
        $\mathbf{I'} = \text{Dec}[\text{Enc}(\mathbf{I}\odot \mathbf{M})]$\\
        $\hat{\mathbf{I}} = \mathbf{I'} \odot (1-\mathbf{M}) + \mathbf{I} \odot \mathbf{M}$\\
        $\mathbf{S}^{k} = f(\mathbf{I},\hat{\mathbf{I}})$ \hfill\# anomaly score map\\
    }
    $\epsilon_k= \frac{1}{\sum_{i,j} \mathbf{M}_{(i,j)}}{\sum_{x,y}\mathbf{S}^k_{(x,y)}}$ \hfill\# anomaly score\\
    \Return anomaly score $\epsilon_k$, anomaly score map $\mathbf{S}^k$
\end{algorithm}

The corresponding algorithm of the mask refinement is shown in Algorithm~\ref{alg:update}. The mask is updated in the unit of patches to make the algorithm more stable and reduce the number of iterations. Thus, given the grid size $k\in K$, the image $\textbf{I}$ is split into $N_k$ $k\times k$ patches $p_1,p_2,\cdots,p_{N_k}$. Then, for each patch $p$, we calculate the average reconstruction error $\epsilon_{p}$ on $p$ and use $\epsilon_{p}$ as a credential to update the mask $\textbf{M}$ according to a threshold $\eta$. The threshold $\eta$ is set to be the maximum error in the validation set. Since the validation set contains only normal samples, this threshold can be considered as a rough boundary between the normal and anomalous cases. We terminate this approach until $\textbf{M}$ converges. For each grid size $k$, we obtain an anomaly score $\epsilon_k$ and an anomaly score map $\mathbf{S}^k$ according to the mask refinement algorithm.

To obtain the final anomaly score map $\mathbf{S}^{final}$, we average all the anomaly score maps $\mathbf{S}^k$ provided by those using different grid sizes of masks. Similarly, the anomaly score $\epsilon$ is computed as the average of $\epsilon_k$ under several grid sizes $k\in K$.

\renewcommand \arraystretch{1.0}
\begin{table*}[t]
\centering
\caption{Results of \textbf{anomaly detection} in terms of AUC in \% on the Retinal-OCT dataset, comparing with several state-of-the-arts. The best-performing method is in bold.}
\label{tal:oct}
\small
\setlength{\tabcolsep}{0.8pt}{
\begin{tabular}{C{1.6cm}C{1.3cm}C{1.3cm}C{1.5cm}C{1.2cm}C{1.2cm}C{1.0cm}C{1.2cm}C{1.2cm}C{1.2cm}C{1.0cm}C{1.0cm}C{1.0cm}C{1.2cm}}
\toprule
Method & 
\makecell[c]{GeoTrans\\\cite{golan2018deep}} &
\makecell[c]{EGBAD\\\cite{zenati2018efficient}} & \makecell[c]{f-AnoGAN\\\cite{schlegl2019f}} & \makecell[c]{DSVDD\\\cite{ruff2018deep}} &
\makecell[c]{Pix2Pix\\\cite{isola2017image}} &
\makecell[c]{AE\\\cite{zhou2017anomaly}} & \makecell[c]{AnoGan\\\cite{schlegl2017unsupervised}} & \makecell[c]{EnGAN\\\cite{han2020gan}} &
\makecell[c]{C-GAN\\\cite{zhu2017unpaired}} &
\makecell[c]{V-GAN\\\cite{baur2018deep}} & \makecell[c]{P-Net\\\cite{zhou2020encoding}} & \makecell[c]{MKD \\ \cite{MKD}} & \cellcolor{gray!15}\makecell[c]{SSM \\ (ours)}\\

\cmidrule(lr){1-1} \cmidrule(lr){2-2} \cmidrule(lr){3-3} \cmidrule(lr){4-4} \cmidrule(lr){5-5} \cmidrule(lr){6-6} 
\cmidrule(lr){7-7} \cmidrule(lr){8-8} \cmidrule(lr){9-9} \cmidrule(lr){10-10} \cmidrule(lr){11-11} \cmidrule(lr){12-12} \cmidrule(lr){13-13} \cmidrule(lr){14-14} 
AUC ($\%$) & 60.1 & 61.0 & 66.6 & 74.4 & 79.4 & 82.1 & 84.8 & 86.9 & 87.4 & 90.6 & 92.9 & 97.0 & \cellcolor{gray!15}\textbf{98.3}\\
\bottomrule
\end{tabular}}
\end{table*}

\renewcommand \arraystretch{1.0}
\begin{table*}[t]
\centering
\caption{Results of \textbf{anomaly detection} on the MVTec AD dataset. Results are listed as AUC in \% and are marked individually for each class. An average score over all classes is also reported in the last row. Results of GeoTrans, GANomaly and ARNet are borrowed from~\cite{fye2020ARNet}. Results of OCGAN, ALOCC, DAE, MemAE and SCADN are borrowed from~\cite{yan2021learning}. Results of MKD are borrowed from~\cite{MKD}. Results of MemSTC are borrowed from~\cite{zhou2021memorizing}. The best-performing method is in bold.}
\label{tal:mvtec1}
\small
\setlength{\tabcolsep}{0.8pt}{
\begin{tabular}{C{1.7cm}C{1.3cm} C{1.3cm} C{1.3cm} C{1.3cm} C{1.3cm} C{1.5cm}C{1.3cm} C{1.3cm} C{1.2cm} C{1.3cm} C{1.3cm}C{1.3cm}}
\toprule
Category &
\makecell[c]{OCGAN \\ \cite{perera2019ocgan}} &
\makecell[c]{ALOCC\\ \cite{Sabokrou2018Adversarially}} &
\makecell[c]{GeoTrans \\ \cite{golan2018deep}} &
\makecell[c]{DAE\\ \cite{Sakurada2014Anomaly}} &
\makecell[c]{MemAE \\ \cite{gong2019memorizing}} &
\makecell[c]{GANomaly \\ \cite{akccay2019skip}} &
\makecell[c]{SCADN \\ \cite{yan2021learning}} &
\makecell[c]{ARNet \\ \cite{fye2020ARNet}} &
\makecell[c]{US \\ \cite{bergmann2020uninformed}} & \makecell[c]{MKD \\ \cite{MKD}} &
\makecell[c]{MemSTC\\\cite{zhou2021memorizing}} &
\cellcolor{gray!15}\makecell[c]{SSM\\(ours)}\\
\cmidrule(lr){1-1} \cmidrule(lr){2-2} \cmidrule(lr){3-3} \cmidrule(lr){4-4} \cmidrule(lr){5-5} \cmidrule(lr){6-6} \cmidrule(lr){7-7} \cmidrule(lr){8-8} \cmidrule(lr){9-9} \cmidrule(lr){10-10} \cmidrule(lr){11-11} \cmidrule(lr){12-12} \cmidrule(lr){13-13}

Bottle & 59.2 & 46.0 & 74.4 & 86.0 & 93.0 & 89.2 & 95.7 & 94.1 & 99.0 & 99.4 & 97 & \cellcolor{gray!15}\textbf{99.9}\\
Cable & 49.6 & 53.1 & 78.3 & 64.8 & 78.5 & 74.5 & 85.6 & 83.2 & 86.2 & \textbf{89.2} & 81 & \cellcolor{gray!15}77.3\\
Capsule & 71.4 & 48.7 & 67.0 & 53.4 & 73.5 & 73.2 & 76.5 & 68.1 & 86.1 & 80.5 & 87 & \cellcolor{gray!15}\textbf{91.4}\\
Carpet & 34.8 & 42.3 & 43.7 & 58.8 & 38.6 & 69.9 & 50.4 & 70.6 & \textbf{91.6} & 79.3 & 61 & \cellcolor{gray!15}76.3\\
Grid & 85.5 & 78.1 & 61.9 & 85.8 & 80.5 & 70.8 & 98.3 & 88.3 & 81.0 & 78.0 & 99 & \cellcolor{gray!15}\textbf{100}\\
Hazelnut & 75.3 & \textbf{99.3} & 35.9 & 51.3 & 76.9 & 78.5 & 83.3 & 85.5 & 93.1 & 98.4 & 98 & \cellcolor{gray!15}91.5\\
Leather & 62.4 & 76.8 & 84.1 & 49.7 & 42.3 & 84.2 & 65.9 & 86.2 & 88.2 & 95.1 & 87 & \cellcolor{gray!15}\textbf{99.9}\\
Metal Nut & 29.5 & 70.5 & 81.3 & 79.3 & 65.4 & 70.0 & 62.4 & 66.7 & 82.0 & 73.6 & 82 & \cellcolor{gray!15}\textbf{88.7}\\
Pill & 70.2 & 72.6 & 63.0 & 69.3 & 71.7 & 74.3 & 81.4 & 78.6 & 87.9 & 82.7 & 87 & \cellcolor{gray!15}\textbf{89.1}\\
Screw & 50.5 & 99.5 & 50.0 & 71.9 & 25.7 & 74.6 & 83.1 & \textbf{100} & 54.9 & 83.3 & 99 & \cellcolor{gray!15}85.0\\
Tile & 80.6 & 52.6 & 41.7 & 89.4 & 71.8 & 79.4 & 79.2 & 73.5 & \textbf{99.1} & 91.6 & 98 & \cellcolor{gray!15}94.4\\
Toothbrush & 59.4 & 64.2 & 97.2 & 94.2 & 96.7 & 65.3 & 98.1 & \textbf{100} & 95.3 & 92.3 & \textbf{100} & \cellcolor{gray!15}\textbf{100}\\
Transistor & 47.7 & 75.1 & 86.9 & 37.6 & 79.1 & 79.2 & 86.3 & 84.3 & 81.8 & 85.6 & 89 & \cellcolor{gray!15}\textbf{91.0}\\
Wood & 95.9 & 27.9 & 61.1 & 88.2 & 95.4 & 83.4 & 96.8 & 92.3 & 97.7 & 94.3 & \textbf{98} & \cellcolor{gray!15}95.9\\
Zipper & 36.4 & 54.7 & 82.0 & 81.9 & 71.0 & 74.5 & 84.6 & 87.6 & 91.9 & 93.2 & 93 & \cellcolor{gray!15}\textbf{99.9}\\
\cmidrule(lr){1-1} \cmidrule(lr){2-2} \cmidrule(lr){3-3} \cmidrule(lr){4-4} \cmidrule(lr){5-5} \cmidrule(lr){6-6} \cmidrule(lr){7-7} \cmidrule(lr){8-8} \cmidrule(lr){9-9} \cmidrule(lr){10-10} \cmidrule(lr){11-11} \cmidrule(lr){12-12} \cmidrule(lr){13-13}
Mean & 60.6 & 64.1 & 67.2 & 70.7 & 70.7 & 76.2 & 81.8 & 83.9 & 87.7 & 87.7 & 90 & \cellcolor{gray!15}\textbf{92.0}\\
\bottomrule
\end{tabular}}
\end{table*}

\subsection{Discussion}
In our framework, the progressive mask refinement approach is only deployed in the inference stage. A natural question is: Why not use it in the training stage? To answer this question, let us recall the definition of anomaly detection. In the training stage, with only normal data provided, it is meaningless to update the mask: the entire image is expected to be well recovered, ending up with an empty mask after the progressive mask refinement. This also violates the conditional reconstruction principle we proposed. Thus, a better choice is reconstructing the images with randomly generated masks. With a larger exploration space of masks, SSM leads to a more powerful representation learning. This design alleviates the impact of the lack of anomalous data during training. We believe that this is a feasible research avenue for unsupervised anomaly detection and localization.

\section{Experiments}
In this section, our method is applied to the Retinal-OCT dataset~\cite{kermany2018identifying} for unsupervised anomaly detection on medical diagnosis, and the MVTec Anomaly Detection~\cite{bergmann2019mvtec} dataset for both image-level anomaly detection and pixel-level anomaly localization, compared with state-of-the-art methods. To evaluate the effectiveness of our method, we further conduct ablation studies under different loss functions and architecture designs of SSM. Finally, we provide visualization analysis to illustrate the effectiveness of the proposed progressive mask refinement approach.

\subsection{Experimental Setups}
\subsubsection{Evaluation Protocols} 
We quantify the model performance using the area under the Receiver Operating Characteristic (ROC) curve metric (AUC). This evaluation protocol allows comparison using different thresholds on the anomaly score. It is commonly adopted as the performance measurement in anomaly detection tasks.

\subsubsection{Datasets}
We conduct experiments on two real-world anomaly detection datasets, which are related to the medical diagnosis and the industrial defect detection:

\textbf{Retinal-OCT dataset}~\cite{kermany2018identifying} is a recent dataset for detecting abnormalities in retinal optical coherence tomography (OCT) images. It contains 84,495 high-resolution clinical images. It has three small disease classes (CNV, DME, DRUSEN) and a large class of disease-free images. We use the large disease-free class as normal data; then, we use the three disease classes together as a single anomalous class. The training-testing split is set the same as the original dataset.

\textbf{MVTec Anomaly Detection dataset}~\cite{bergmann2019mvtec} comprises 15 categories with 3629 images for training and validation and 1725 images for testing. The training set contains only normal images without defects. The test set contains images containing various kinds of defects and defect-free images. In total, 73 different defect types are present, on average five per category. Five categories cover different types of regular (carpet, grid) or random (leather, tile, wood) textures, while the remaining ten categories represent various types of objects. All image resolutions are in the range between $700 \times 700$ and $1024 \times 1024$ pixels. Pixel-precise ground truth labels for each defective image region are provided. The dataset contains almost 1900 manually annotated regions.

\renewcommand \arraystretch{1.0}
\begin{table*}[t]
\centering
\caption{Results of \textbf{anomaly localization} on the MVTec AD dataset. Results are listed as AUC in \% and are marked individually for each class. An average score over all classes is also reported in the last row. Results of DAE, OCGAN, MemAE and SCADN are borrowed from~\cite{yan2021learning}. Results of AnoGAN, CNN-dict and SMAI are borrowed from~\cite{li2020superpixel}. Results of GDR and MKD are borrowed from~\cite{MKD}. Results of GP are borrowed from~\cite{GP}. The best-performing method is in bold.}
\label{tal:mvtec2}
\small
\setlength{\tabcolsep}{0.8pt}{
\begin{tabular}{C{1.7cm}C{1.3cm}C{1.3cm}C{1.3cm}C{1.3cm}C{1.3cm}C{1.3cm}C{1.3cm}C{1.3cm}C{1.3cm}C{1.0cm}C{1.3cm}C{1.3cm}}
\toprule
Category & \makecell[c]{DAE\\ \cite{Sakurada2014Anomaly}} & \makecell[c]{OCGAN \\ \cite{perera2019ocgan}}
& \makecell[c]{MemAE \\ \cite{gong2019memorizing}} & \makecell[c]{AnoGAN \\ \cite{schlegl2017unsupervised}} & \makecell[c]{SCADN \\ \cite{yan2021learning}} &
\makecell[c]{CNN-dict \\ \cite{napoletano2018anomaly}} & \makecell[c]{SMAI\\\cite{li2020superpixel}} &
\makecell[c]{GDR \\ \cite{dehaene2020iterative}} &
\makecell[c]{MemSTC\\\cite{zhou2021memorizing}} &
\makecell[c]{MKD \\ \cite{MKD}} & \makecell[c]{GP\\ \cite{GP}} &
\cellcolor{gray!15}\makecell[c]{SSM\\(ours)} \\
\cmidrule(lr){1-1} \cmidrule(lr){2-2} \cmidrule(lr){3-3} \cmidrule(lr){4-4} \cmidrule(lr){5-5} \cmidrule(lr){6-6} \cmidrule(lr){7-7} \cmidrule(lr){8-8} \cmidrule(lr){9-9} \cmidrule(lr){10-10} \cmidrule(lr){11-11} \cmidrule(lr){12-12} \cmidrule(lr){13-13}

Bottle & 54.4 & 56.7 & 72.4 & 86 & 69.6 & 78 & 86 & 92.2 & 87.2 & \textbf{96.3} & 93 & \cellcolor{gray!15}95.9\\
Cable & 53.5 & 56.4 & 81.4 & 78 & 81.4 & 79 & 92 & 91.0 & 91.2 & 82.4 & \textbf{94} & \cellcolor{gray!15}82.1\\
Capsule  & 54.2 & 63.7 & 67.3 & 84 & 68.7 & 84 & 93 & 91.7 & 91.2 & 95.9 & 90 & \cellcolor{gray!15}\textbf{98.4}\\
Carpet & 52.8 & 54.6 & 57.4 & 54 & 64.9 & 72 & 88 & 73.5 & 85.7 & 95.6 & \textbf{96} & \cellcolor{gray!15}94.4\\
Grid & 55.0 & 65.2 & 46.8 & 58 & 79.6 & 59 & 97 & 96.1 & 93.9 & 91.8 & 78 & \cellcolor{gray!15}\textbf{99.0}\\
Hazelnut & 66.4 & 84.1 & 84.6 & 87 & 88.4 & 72 & 97 & \textbf{97.6} & 96.1 & 94.6 & 84 & \cellcolor{gray!15}97.4\\
Leather & 78.3 & 74.9 & 68.6 & 64 & 76.3 & 87 & 86 & 92.5 & 95.7 & 98.1 & 90 & \cellcolor{gray!15}\textbf{99.6}\\
Metal Nut & 53.9 & 53.4 & 76.9 & 76 & 75.4 & 82 & \textbf{92} & 90.7 & 89.0 & 86.4 & 91 & \cellcolor{gray!15}89.6\\
Pill & 55.5 & 59.6 & 73.7 & 87 & 74.7 & 68 & 92 & 93.0 & 93.1 & 89.6 & 93 & \cellcolor{gray!15}\textbf{97.8}\\
Screw & 57.0 & 70.8 & 73.2 & 80 & 87.6 & 87 & 96 & 94.5 & 90.1 & 96.0 & 96 & \cellcolor{gray!15}\textbf{98.9}\\
Tile & 63.0 & 59.2 & 64.7 & 50 & 67.7 & \textbf{93} & 62 & 65.4 & 85.9 & 82.8 & 80 & \cellcolor{gray!15}90.2\\
Toothbrush & 61.6 & 76.3 & 88.6 & 90 & 90.1 & 77 & 96 & 98.5 & 95.2 & 96.1 & 96 & \cellcolor{gray!15}\textbf{98.9}\\
Transistor & 53.2 & 58.2 & 71.4 & 80 & 68.9 & 66 & 85 & 91.9 & 86.9 & 76.5 & \textbf{100} & \cellcolor{gray!15}80.1\\
Wood & 61.2 & 65.5 & 65.2 & 62 & 67.2 & \textbf{91} & 80 & 83.8 & 85.1 & 84.8 & 81 & \cellcolor{gray!15}86.9\\
Zipper & 53.6 & 62.4 & 64.3 & 78 & 67.0 & 76 & 90 & 86.9 & 89.4 & 93.9 & \textbf{99} & \cellcolor{gray!15}\textbf{99.0}\\
\cmidrule(lr){1-1} \cmidrule(lr){2-2} \cmidrule(lr){3-3} \cmidrule(lr){4-4} \cmidrule(lr){5-5} \cmidrule(lr){6-6}  \cmidrule(lr){7-7} \cmidrule(lr){8-8} \cmidrule(lr){9-9} \cmidrule(lr){10-10} \cmidrule(lr){11-11} \cmidrule(lr){12-12} \cmidrule(lr){13-13}
Mean & 58.2 & 64.1 & 70.4 & 74 & 75.2 & 78 & 89 & 89.3 & 90.4 & 90.7 & 91 & \cellcolor{gray!15}\textbf{93.9}\\
\bottomrule
\end{tabular}}
\end{table*}

\subsubsection{Baselines}

For anomaly detection, we consider several state-of-the-art methods as baselines. For the one-class classification-based method, we consider DSVDD~\cite{ruff2018deep} as the baseline. For the reconstruction-based mathods, DAE~\cite{Sakurada2014Anomaly}, AE~\cite{zhou2017anomaly}, MemAE~\cite{gong2019memorizing}, SCADN~\cite{yan2021learning} and MemSTC~\cite{zhou2021memorizing} are considered. For GAN-based methods, Pix2Pix~\cite{isola2017image}, AnoGAN~\cite{schlegl2017unsupervised}, C-GAN~\cite{zhu2017unpaired}, V-GAN~\cite{baur2018deep}, EGBAD~\cite{zenati2018efficient}, ALOCC~\cite{Sabokrou2018Adversarially}, GANomaly~\cite{akccay2019skip}, f-AnoGAN~\cite{schlegl2019f}, OCGAN~\cite{perera2019ocgan} and EnGAN~\cite{han2020gan} are considered. For other self-supervision-based methods, we consider GeoTrans~\cite{golan2018deep} and ARNet~\cite{fye2020ARNet}. We also consider the knowledge distillation-based methods, US~\cite{bergmann2020uninformed} and MKD~\cite{MKD}, which use additional data to pre-train the networks, to establish strong baselines. Especially, in medical diagnosis, we consider P-Net~\cite{zhou2020encoding} as a baseline, which uses additional medical domain knowledge and uses a large amount of additional data and annotations for the training. 

For anomaly localization, we follow~\cite{li2020superpixel,MKD,GP} and consider DAE~\cite{Sakurada2014Anomaly}, OCGAN~\cite{perera2019ocgan}, MemAE~\cite{gong2019memorizing}, AnoGAN~\cite{schlegl2017unsupervised}, SCADN~\cite{yan2021learning}, CNN-dict~\cite{napoletano2018anomaly}, SMAI~\cite{li2020superpixel}, GDR~\cite{dehaene2020iterative}, MKD~\cite{MKD} and GP~\cite{GP} as baselines.

\subsubsection{Model Configuration}

For the encoder-decoder structure for SSM, we follow the settings in~\cite{unet,fye2020ARNet,zhou2020encoding,li2020superpixel} and add skip-connections between some layers in the encoder and corresponding decoder layers to facilitate the backpropagation of the gradient in an attempt to improve the performance of image reconstruction. The individual loss weights $\lambda_1$, $\lambda_2$, $\lambda_3$ and $\lambda_4$ are set to 1 as default. We use Adam~\cite{kingma2014adam} optimizer with the weight decay of $1e^{-5}$. Other hyperparameters are default in Pytorch. SSM is trained using a batch size of 8 for 300 epochs with one NVIDIA GTX 3090. The learning rate is initially set to $1e^{-4}$, and is divided by 2 every 50 epochs.

\subsection{Comparison with State-of-the-art Methods}
In this section, we show quantitative results of the proposed SSM, comparing with several state-of-the-art methods on the Retinal-OCT dataset and MVTec AD dataset.

\textbf{Results on Retinal-OCT dataset.} Table~\ref{tal:oct} shows the AUC results of anomaly detection on Retinal-OCT dataset, comparing with several state-of-the-arts, including GeoTrans~\cite{golan2018deep}, EGBAD~\cite{zenati2018efficient}, f-AnoGAN~\cite{schlegl2019f}, DSVDD~\cite{ruff2018deep}, Pix2Pix~\cite{isola2017image}, AE~\cite{zhou2017anomaly},  AnoGAN~\cite{schlegl2017unsupervised}, EnGAN~\cite{han2020gan}, C-GAN~\cite{zhu2017unpaired}, V-GAN~\cite{baur2018deep}, P-Net~\cite{zhou2020encoding} and MKD~\cite{MKD}. Results show that on the Retinal-OCT dataset, the proposed SSM outperforms all the state-of-the-art methods, especially the two methods, P-Net and MKD, that utilize a large amount of additional data for the training. In detail, P-Net is a special method for detecting the anomaly in retinal images. It uses structural information of retinal images, which is extracted by a segmentation network pre-training with a large amount of additional data and pixel-level annotations~\cite{zhou2020encoding}. The knowledge distillation-based methods MKD uses ImageNet~\cite{russakovsky2015imagenet} to pre-train the networks and thus obtains powerful feature representation abilities. The proposed SSM uses only the data in the training set of the Retinal-OCT dataset, without any other additional data. Under this setting, SSM still shows better performance (98.3\% AUC) than P-Net (92.9\% AUC) and MKD (97.0\% AUC), showing the effectiveness of the proposed method.

\begin{figure}[t]
    \centering
    \includegraphics[width=0.5\textwidth]{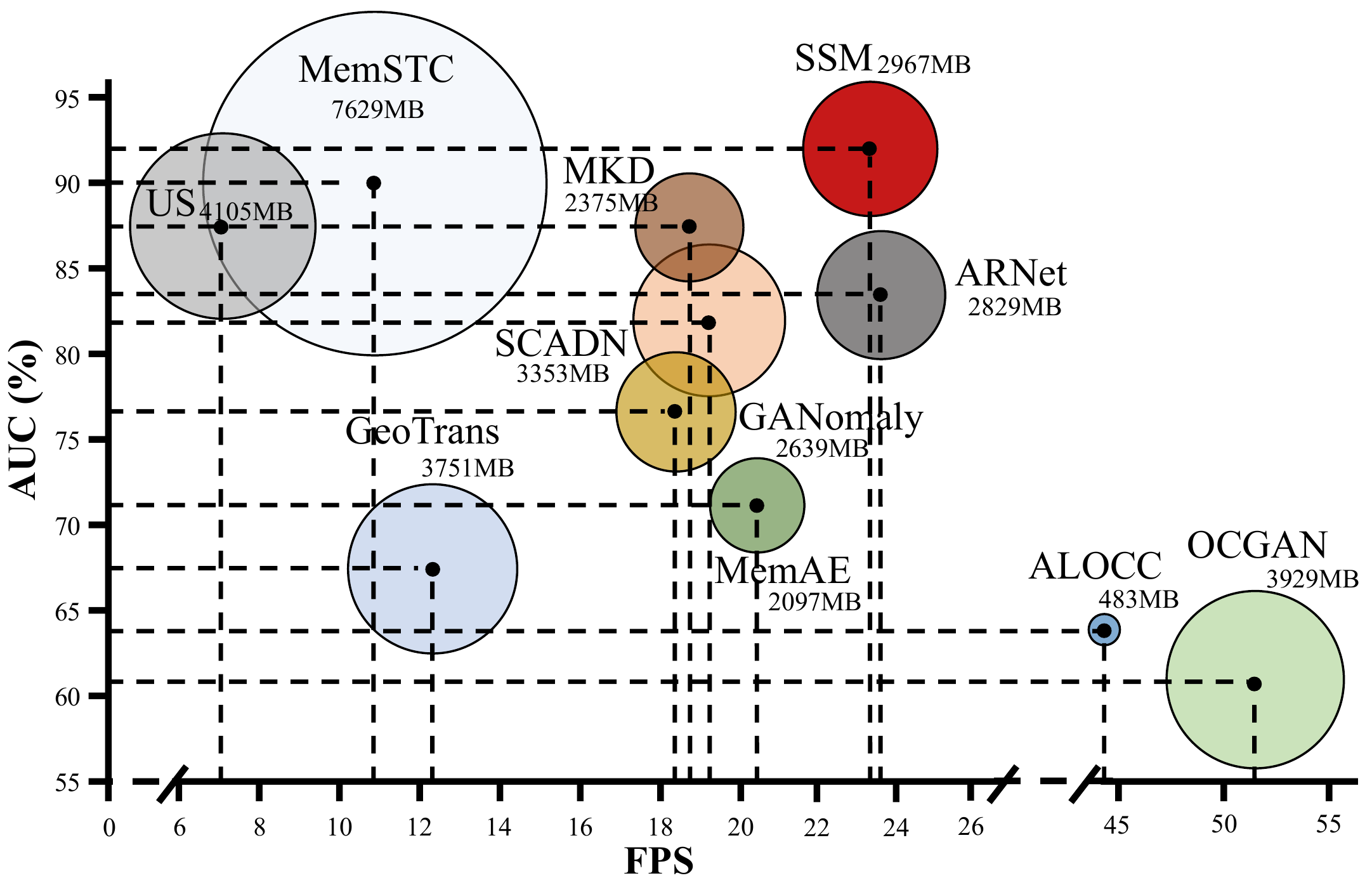}
    \caption{Comparison of frames per second (FPS) (horizontal coordinates), GPU memory usages (circular sizes) and AUC for anomaly detection (vertical coordinates) of various methods testing on MVTec. The proposed SSM takes up a relatively small GPU memory, and its FPS is relatively higher.}
    \label{fig:speed}
\end{figure}

\renewcommand \arraystretch{1.0}
\begin{table}[t]
\centering
\caption{Ablation studies on different losses and architectures of SSM. Results are shown as AUC in \% on the Retinal-OCT dataset. The best-performing method is in bold.}
\label{tal:ablation}
\setlength{\tabcolsep}{1.5pt}{
\begin{tabular}{C{1.0cm}C{1.0cm}C{1.0cm}C{1.0cm}|C{1.0cm}C{1.4cm}|C{1.6cm}}
\toprule
\multicolumn{4}{c}{Losses} & \multicolumn{2}{c}{Architectures} & \\
$\mathcal{L}_{\textit{MSE}}$ &
$\mathcal{L}_{\textit{SSIM}}$ & 
$\mathcal{L}_{\textit{GMS}}$ &
$\mathcal{L}_{\textit{MASK}}$ & 
MAM & Refinement & AUC ($\%$) \\
\cmidrule(lr){1-1} \cmidrule(lr){2-2} \cmidrule(lr){3-3} \cmidrule(lr){4-4} \cmidrule(lr){5-5} \cmidrule(lr){6-6} \cmidrule(lr){7-7}
\checkmark & & & & & & 94.3\\
\checkmark & & & \checkmark & & & 94.7\\
&\checkmark& &\checkmark & & & 74.8\\
& &\checkmark&\checkmark & & & 94.0\\
\cmidrule(lr){1-1} \cmidrule(lr){2-2} \cmidrule(lr){3-3} \cmidrule(lr){4-4} \cmidrule(lr){5-5} \cmidrule(lr){6-6} \cmidrule(lr){7-7}
\checkmark &\checkmark&\checkmark&\checkmark& & & 96.2\\
\checkmark &\checkmark&\checkmark &\checkmark& \checkmark & & 96.8\\
\checkmark &\checkmark&\checkmark& \checkmark & & \checkmark & 97.6\\
\cellcolor{gray!15}\checkmark &\cellcolor{gray!15}\checkmark&\cellcolor{gray!15}\checkmark & \cellcolor{gray!15}\checkmark& \cellcolor{gray!15}\checkmark & \cellcolor{gray!15}\checkmark &\cellcolor{gray!15} \textbf{98.3}\\
\bottomrule
\end{tabular}}
\end{table}

\renewcommand \arraystretch{1.0}
\begin{table}[t]
\centering
\caption{Ablation studies for anomaly detection and localization on different grid sizes of masks for SSM. Results are shown as AUC in \%. The best-performing method is in bold.}
\label{tal:ablation2}
\setlength{\tabcolsep}{1.3pt}{
\begin{tabular}{C{0.75cm}C{0.75cm}C{0.75cm}C{0.75cm}|C{1.7cm}C{1.7cm}|C{1.7cm}}
\toprule
\multicolumn{4}{c}{Grid Sizes of Masks} & \multicolumn{2}{c}{Detection} & Localization\\
4 & 8 & 16 & 32 & 
Retinal-OCT & MVTec AD & MVTec AD
\\
\cmidrule(lr){1-1} \cmidrule(lr){2-2} \cmidrule(lr){3-3} \cmidrule(lr){4-4} \cmidrule(lr){5-5} \cmidrule(lr){6-6} \cmidrule(lr){7-7}
\checkmark & \checkmark &  &  & 96.8 & 90.8 & 93.2\\
& \checkmark & \checkmark &  & 98.0 & 91.4 & 93.4\\
&  & \checkmark & \checkmark & 94.3 & 91.3 & 93.1\\
\cellcolor{gray!15}\checkmark & \cellcolor{gray!15}\checkmark & \cellcolor{gray!15}\checkmark & \cellcolor{gray!15} & \cellcolor{gray!15}\textbf{98.3} & \cellcolor{gray!15}\textbf{92.0} & \cellcolor{gray!15}\textbf{93.9}\\
& \checkmark & \checkmark & \checkmark & 94.6 & 91.4 & 93.2\\
\checkmark & \checkmark & \checkmark & \checkmark & 94.9 & 91.9 & 93.8 \\
\bottomrule
\end{tabular}}
\end{table}

\begin{figure}[t]
    \centering
    \includegraphics[width=0.38\textwidth]{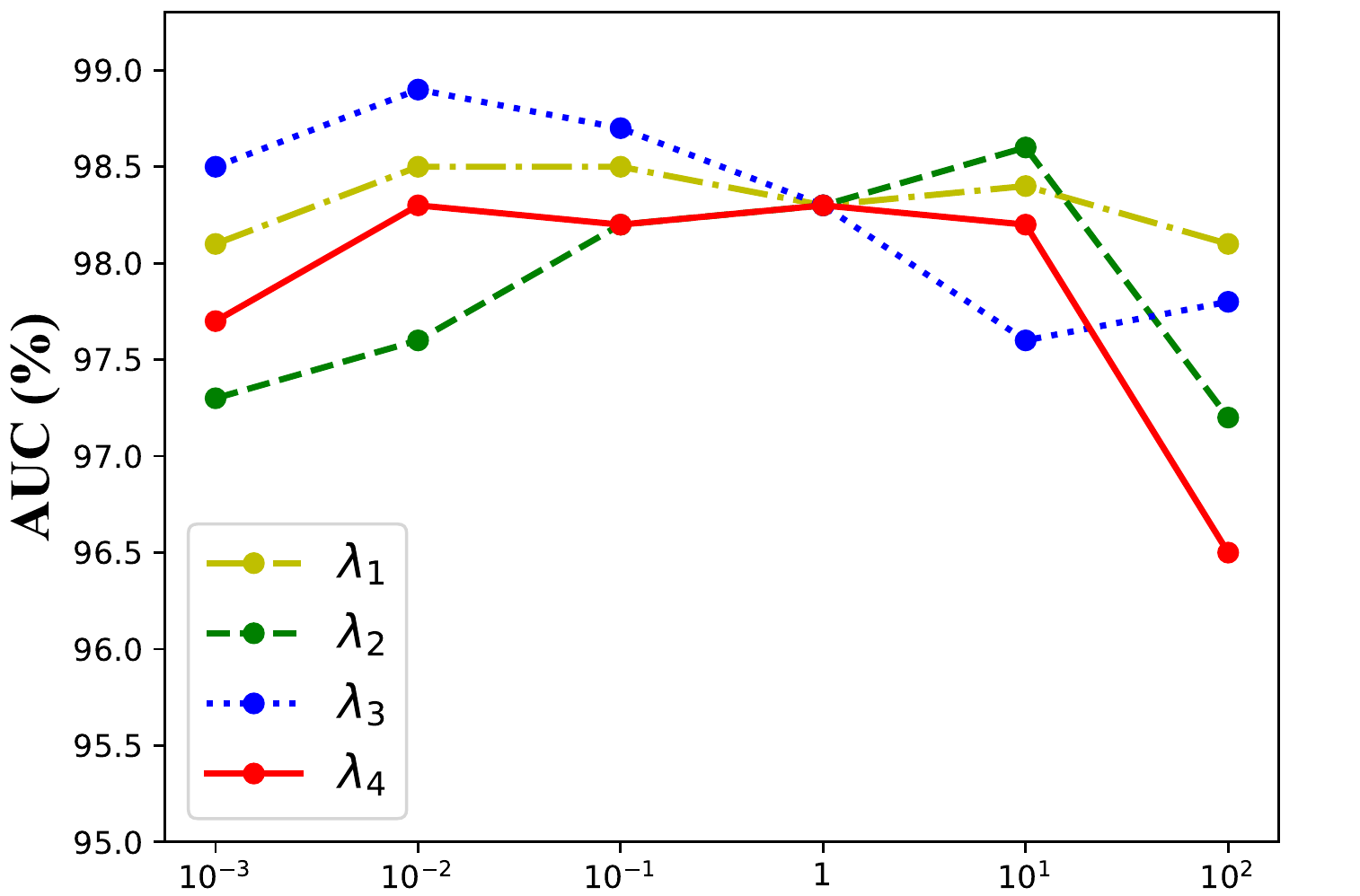}
    \caption{Sensitivity analysis w.r.t. the hyperparameters for SSM on Retinal-OCT. The AUC for anomaly detection is reported. Best viewed in color.}
    \label{fig:lambda}
\end{figure}

\textbf{Results on MVTec AD dataset.} For the task of anomaly detection on the MVTec AD dataset, Table~\ref{tal:mvtec1} shows the corresponding results. We compare the proposed SSM with several state-of-the-arts, including OCGAN~\cite{perera2019ocgan}, ALOCC~\cite{Sabokrou2018Adversarially}, GeoTrans~\cite{golan2018deep}, DAE~\cite{Sakurada2014Anomaly}, MemAE~\cite{gong2019memorizing}, GANomaly~\cite{akccay2019skip}, SCADN~\cite{yan2021learning}, ARNet~\cite{fye2020ARNet}, US~\cite{bergmann2020uninformed} and MKD~\cite{MKD}. As shown in Table~\ref{tal:mvtec1}, the proposed SSM achieves the highest mean AUC among all categories (92.0\% AUC, 2.0\% higher than MemSTC). In 9 out of the 15 categories, SSM outperforms all the other baseline methods. For the other 6 categories, the best performance is achieved by 5 different methods. SSM also achieves the least standard deviation (7.88) for the 15 categories, compared to US (10.92), MKD (7.94), and MemSTC (10.43), which shows that SSM has a good generalizability across different categories.

For the task of anomaly localization, we consider DAE~\cite{Sakurada2014Anomaly}, OCGAN~\cite{perera2019ocgan}, MemAE~\cite{gong2019memorizing}, AnoGAN~\cite{schlegl2017unsupervised}, SCADN~\cite{yan2021learning}, CNN-dict~\cite{napoletano2018anomaly}, SMAI~\cite{li2020superpixel}, GDR~\cite{dehaene2020iterative}, MKD~\cite{MKD} and GP~\cite{GP} as the baselines. The corresponding results are shown in Table~\ref{tal:mvtec2}. Overall, results in Table~\ref{tal:mvtec2} show that the proposed SSM outperforms other methods on 7 categories and achieves the highest mean AUC among all categories (93.9\% AUC, 2.9\% higher than GP), showing the effectiveness of anomaly localization of SSM.

\textbf{Limitation}. 
Anomaly regions of different categories vary significantly in attributes such as shape and size. As these attributes are unknown at training, some important hyper-parameters, \emph{e.g.}, the mask grid size, cannot be well determined by prior. As a result, we choose to use the same hyper-parameter setting for all categories, which may not guarantee the best performance for every category at the same time. However, results have shown that the proposed SSM is able to yield the best overall performance on all the categories as it achieves the highest mean AUC compared with all the state-of-the-art methods.

\textbf{Computational Cost}. We investigate the computational efficiency and the cost of GPU memory for SSM, as well as several state-of-the-art methods. The results are shown in Figure~\ref{fig:speed}. For all methods, we test 5 times on the MVTec AD dataset with NVIDIA GTX 3090 and record the average FPS and the GPU memory costs. Among all the methods, OCGAN~\cite{perera2019ocgan} has an advantage in the highest computational efficiency (51.3 fps) but takes up relatively large GPU memory (3929MB). The most state-of-the-art method MemSTC~\cite{zhou2021memorizing} consumes more GPU memory (7629MB) and suffers from a relatively low computational efficiency (10.9 fps). SSM reaches 23.5 fps and takes only 2967MB of GPU memory. Without the mask attention module, SSM reaches a slightly higher speed (24.2fps) and consumes lower GPU memory (2803M). Though during the inference phase, SSM needs to refine the masks with multiple image reconstructions, SSM still reaches a considerable efficiency thanks to its light network structure. Compared with SCADN~\cite{yan2021learning}, the most recent state-of-the-art image inpainting-based method, SSM has a much lighter network structure and traverses fewer possible masks during inference. To summarize, with the best performance in AUC, SSM takes up a relatively small GPU memory with high computational efficiency.

\begin{figure*}[t]
    \centering
    \includegraphics[width=0.98\textwidth]{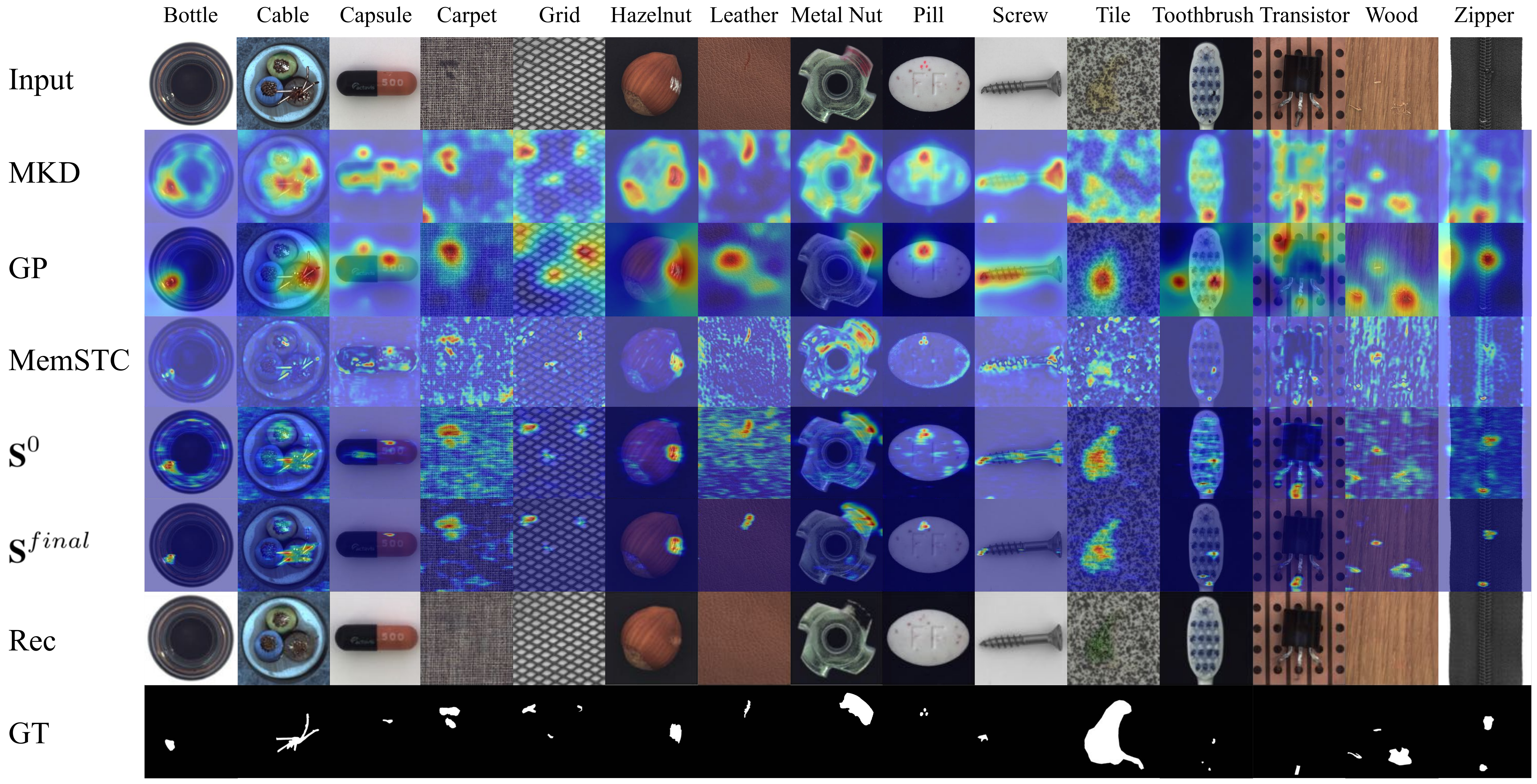}
    \caption{Qualitative results of anomaly localization of SSM on the MVTec AD dataset for several \textbf{difficult cases}, compared with several state-of-the-art methods, including MKD~\cite{MKD}, GP~\cite{GP}, and MemSTC~\cite{zhou2021memorizing}. For SSM, $\mathbf{S}^0$ is the anomaly score map at the initialization step.  $\mathbf{S}^{final}$ is the final anomaly score map after mask refinement by SSM. Rec is the final reconstructed image by SSM. GT is the ground truth.}
    \vspace{-8pt}
    \label{fig:mvtec}
\end{figure*}

\subsection{Ablation Studies}
We study the contribution of the proposed components of SSM independently.

\subsubsection{Loss Functions} 
Table~\ref{tal:ablation} shows experimental results of ablation studies on Retinal-OCT dataset. We first evaluate the impact of the mask reconstruction loss, $\mathcal{L}_{\textit{MASK}}$. As $\mathcal{L}_{\textit{MASK}}$ cannot be use alone to train the model, we compare $\mathcal{L}_{\textit{MASK}}$+$\mathcal{L}_{\textit{MSE}}$ with $\mathcal{L}_{\textit{MSE}}$ alone, and show that adding the mask reconstruction loss leads to an increase in AUC from 94.3\% to 94.7\%. We then investigate the influences of the three image reconstruction losses, $\mathcal{L}_{\textit{MSE}}$, $\mathcal{L}_{\textit{SSIM}}$ and $\mathcal{L}_{\textit{GMS}}$. For this set of experiments, $\mathcal{L}_{\textit{MASK}}$ are always used as default. Among the three losses, $\mathcal{L}_{\textit{MSE}}$ focuses on the error of every pixel in the image and leads to the highest AUC (94.7\%); $\mathcal{L}_{\textit{GMS}}$, a similarity loss function based on image gradient magnitude, is good at detecting anomalous regions resulted from the roughness and bulge of the object edges (94.0\% in AUC); $\mathcal{L}_{\textit{SSIM}}$, focusing on measuring the structural similarity of images, despite its lowest AUC among the three (74.8\%), is good at detecting structural anomalies. Combining the three image reconstruction losses, together with the mask reconstruction loss, the AUC of anomaly detection can be further improved to 96.2\%.

We also note that the baseline method with only the MSE loss has higher AUC (94.3\% as shown in Table~\ref{tal:ablation}) compared to P-Net~\cite{zhou2020encoding} (92.9\% as shown in Table~\ref{tal:oct}), the state-of-the-art reconstruction-based method for anomaly detection, despite the exactly same autoencoder architectures employed. 
The gain in AUC for the baseline method suggests that the proposed random masking and restoring framework, a form of image inpainting, itself outperforms the whole image reconstruction-based framework which P-Net adopts. A similar result is shown in Table~\ref{tal:mvtec1}, where SCADN~\cite{yan2021learning} (a simple image inpainting framework) outperforms the state-of-the-art image reconstruction-based method MemAE~\cite{gong2019memorizing} for $>10\%$ AUC on MVTec AD dataset.

\subsubsection{Architectures} 
The two main important designs of SSM are the mask attention module (MAM) and the progressive mask refinement approach. As one of the self-supervised learning methods, MAM is used to improve semantic feature learning and the representation ability of the latent features. Table~\ref{tal:ablation} shows that the mask attention module (MAM) can steadily improve the anomaly detection performance (from 96.2\% to 96.8\% in AUC). Then, the progressive mask refinement approach (see `Refinement' in Table~\ref{tal:ablation}) can also be used to further improve the anomaly detection performance (from 96.8\% to 98.3\% in AUC). The higher the value is, the more difficult it is to improve the AUC, which shows that the proposed progressive mask refinement approach is highly effective. Since the anomaly detection approach of SSM often contains 2-4 iterations, to build a strong baseline, experiments without the progressive mask refinement approach are conducted on reconstruction tasks under 16 randomly generated masks for one test data.  
\subsubsection{Grid Sizes of Masks} We discuss the basic grid sizes of the masks used for the model training and the mask initialization during the test. Table~\ref{tal:ablation2} shows the corresponding experimental results on Retinal-OCT and MVTec AD dataset. Results show that with $4\times 4$, $8\times 8$ and $16\times 16$ masks, SSM shows the best anomaly detection and localization performance. For example, we achieve 98.3\% AUC for anomaly detection on Retinal-OCT dataset, 92.0\% for anomaly detection and 93.9\% for anomaly localization on MVTec AD dataset. Experiments containing $32\times 32$ masks obtain relatively lower AUC, because too large masks greatly increase the difficulty of image reconstruction. We thus use the mask with sizes of $4\times 4$, $8\times 8$ and $16\times 16$ in the following experiments with default. Complete results for each category on MVTec AD are shown in the appendix.

\subsubsection{Sensitivity Analysis} 
In the previous submission, all experiments are performed under the default setting for SSM, \emph{i.e.}, all individual loss weights $\lambda_i$ ($i=\{1,\cdots, 4\}$) are set to 1. To further study the impact of the weights, we perform a sensitivity analysis for each of the above hyperparameters (fixing the other hyperparameters to 1). As shown in Figure~\ref{fig:lambda}, the performance of SSM is not too much sensitive to the choice of the hyperparameters. Although carefully tuning the hyperparameters may be able to lead to a slighter better result, it may also pose the risk of overfitting and also introduce a lot of computational costs. As a result, we leave the hyperparameters to default.

\begin{figure}[t]
\centering
\begin{minipage}[t]{0.24\textwidth}
\centering
\includegraphics[width=4.3cm]{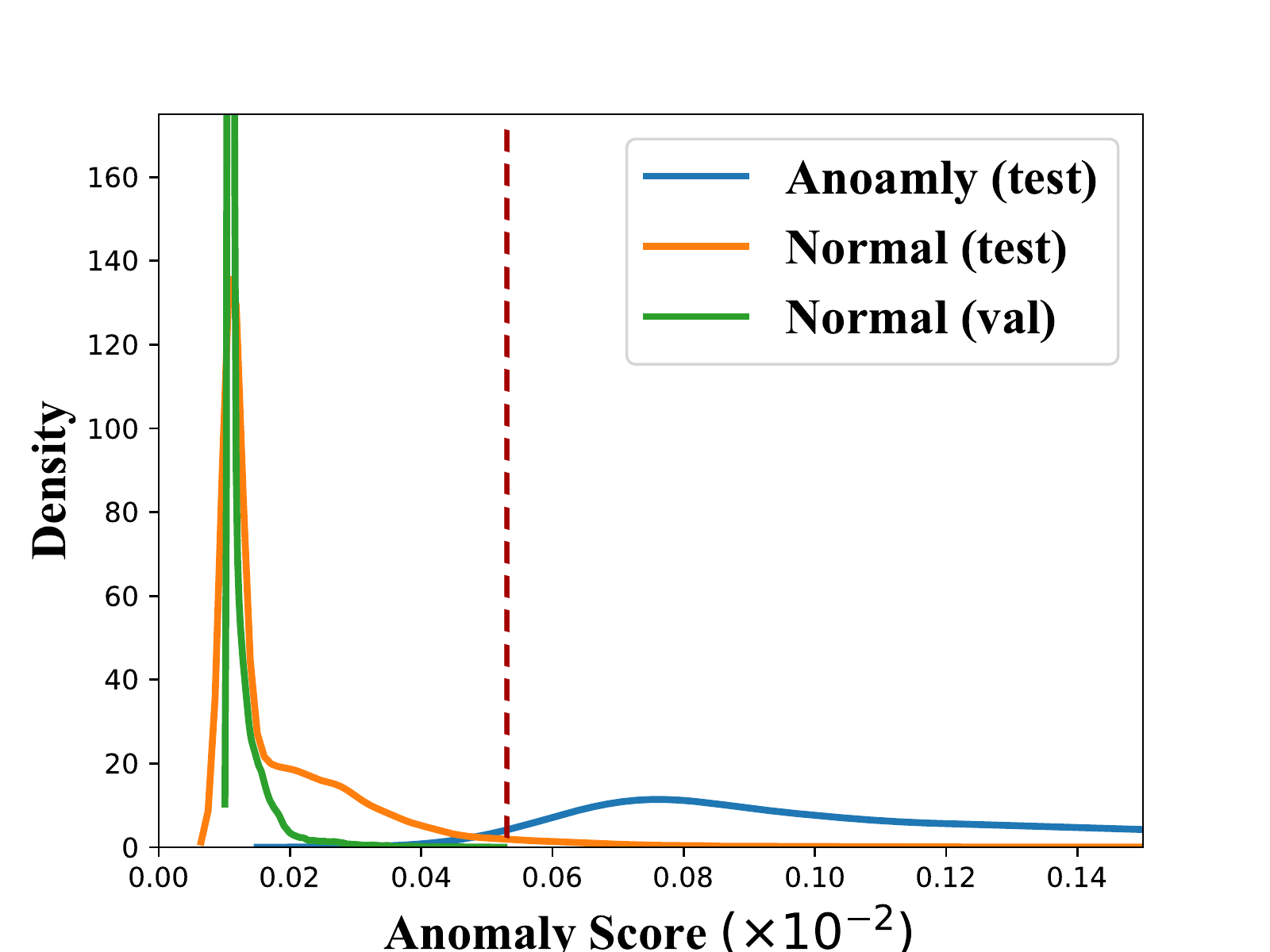}\\
(a) Bottle
\end{minipage}
\begin{minipage}[t]{0.24\textwidth}
\centering
\includegraphics[width=4.3cm]{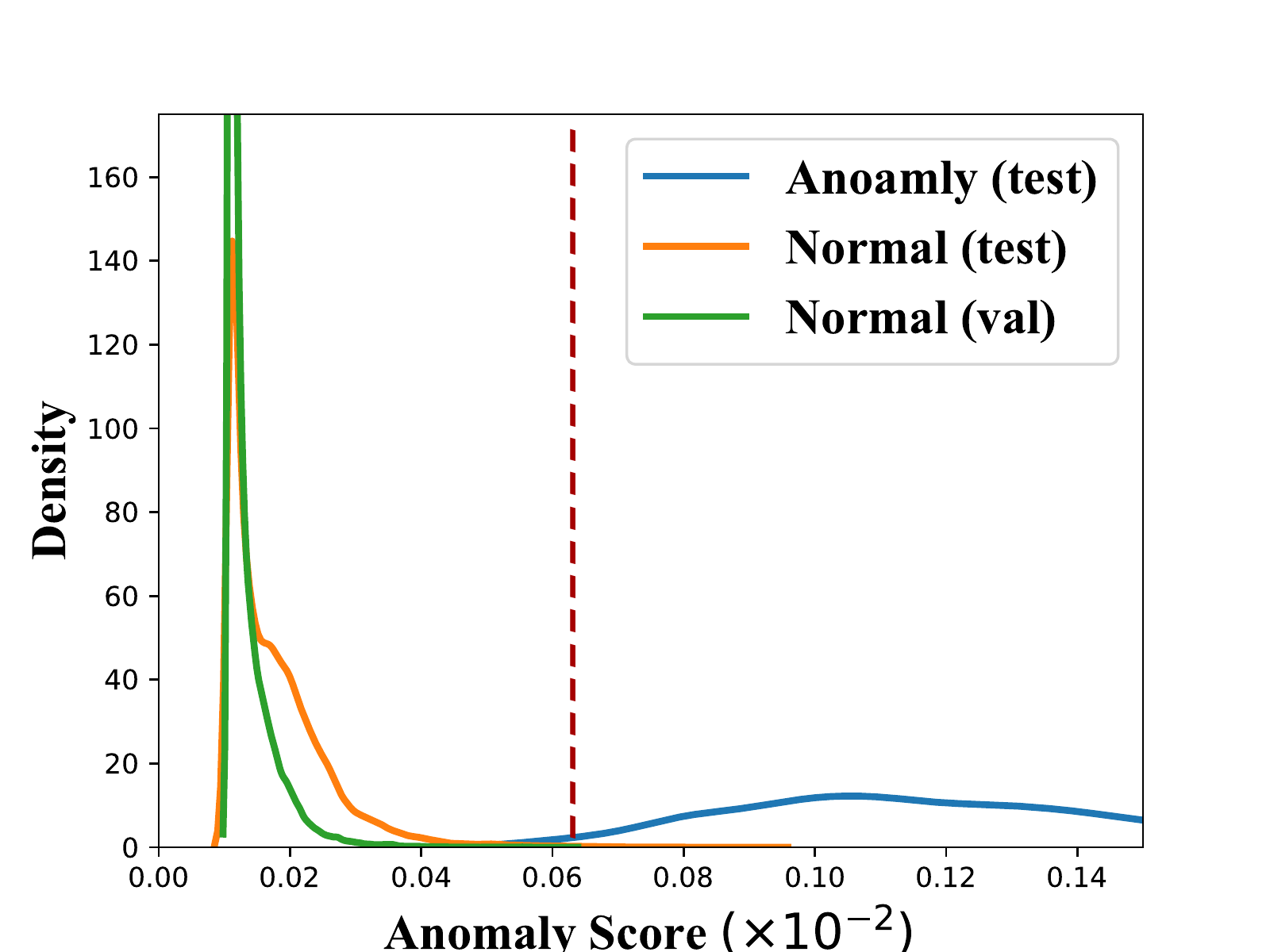}\\
(b) Grid
\end{minipage}
\caption{The pixel-level anomaly score distributions of (a) bottle and (b) grid on the MVTec AD dataset. The red dotted lines represent the threshold utilized in the mask refinement approach, which is set as the maximum anomaly score in the validation set for each category.}
\label{fig:threshold}
\end{figure}

\begin{figure}[t]
    \centering
    \includegraphics[width=0.48\textwidth]{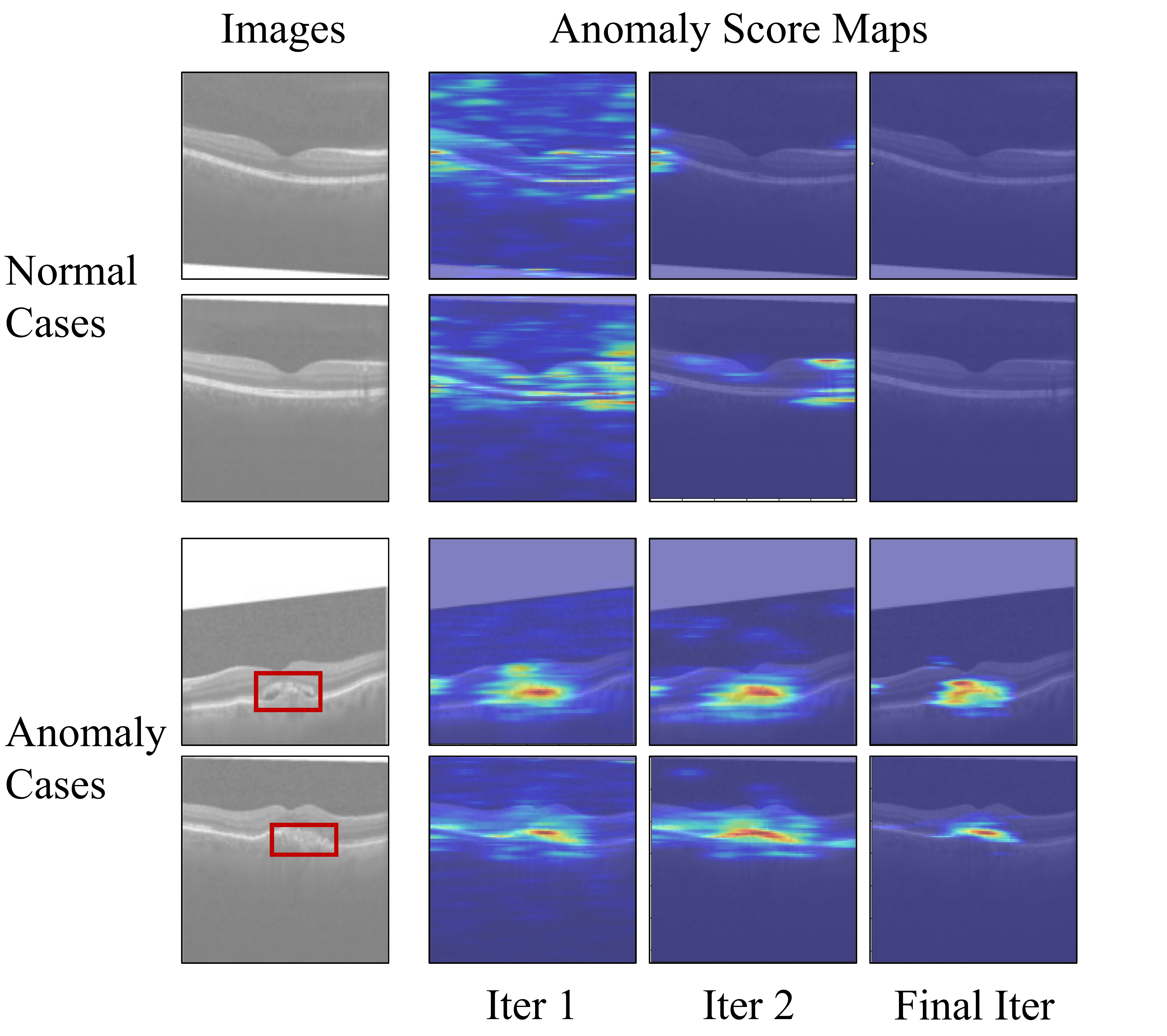}
    \caption{Qualitative results of anomaly localization with SSM on the Retinal-OCT dataset. During the process of mask refinement, SSM continually narrows the scopes of attention maps to only the anomalous regions.}
    \label{fig:retinal}
\end{figure}

\subsection{Visualization Analysis}
To analyze how conditional autoencoder and the corresponding progressive mask refinement approach improve the anomaly localization performance, we visualize the results of some hard cases in the MVTec AD dataset and Retinal-OCT dataset to provide qualitative analysis.

\textbf{MVTec AD dataset.} Figure~\ref{fig:mvtec} shows the visualization of the results from the MVTec AD dataset, including one case for each of the 15 categories, compared with several state-of-the-art methods, including MKD~\cite{MKD}, GP~\cite{GP}, and MemSTC~\cite{zhou2021memorizing}. For SSM, the initialized anomaly score map $\mathbf{S}^0$ is wrongly highlighted in many normal regions. Due to some disturbances such as the light intensity, the shadow, or the special placing angle of the object, the model locates the anomaly in part of normal areas after the initialization phase. Fortunately, after the mask refinement, as shown in $\mathbf{S}^{final}$, SSM narrows the scope of attention for anomaly cases in Figure~\ref{fig:mvtec} and finally successfully located the anomalies, which performs significantly better than the other state-of-the-art methods. The visual analysis strongly illustrates the effectiveness of the progressive mask refinement approach.

We also visualize the anomaly score distributions of bottle and grid on MVTec AD dataset in Figure~\ref{fig:threshold}. The red dotted lines represent the threshold utilized in the progressive mask refinement approach. The values of the thresholds are set as the maximum anomaly score in the validation set. We can clearly see that the normal and anomalous pixels can be well separated by the thresholds.

\textbf{Retinal-OCT dataset.}
We show the visualization results for normal and anomaly cases in Retinal-OCT dataset in Figure~\ref{fig:retinal}. During the process of mask refinement, SSM continually narrows the scopes of attention maps for both normal and anomaly cases. Especially for the anomaly cases, SSM obtains better predictions in the final iteration, which is matched with the related lesion areas (red bounding in the images). For the normal cases, although SSM focuses on some normal areas wrongly after the initialization step (Iter 1), these areas are eventually be cleared, which helps achieve the correct conclusion (Final Iter). The visual analysis well illustrates the effectiveness of the progressive mask refinement approach. 

\section{Conclusion and Future Work}
This paper proposes a novel technique named \emph{\underline{S}elf-\underline{s}upervised \underline{M}asking (SSM)} for unsupervised anomaly detection and localization. A conditional autoencoder is leveraged to learn powerful representations under a larger search space of randomly generated masks in the manner of self-supervised learning. Then we introduce a progressive mask refinement approach to progressively uncover the normal regions and finally locate the anomalous regions. The proposed SSM outperforms state-of-the-arts on multiple anomaly detection benchmarks for both anomaly detection and anomaly localization. Notably, there are still more effective mask refinement approaches to be explored. It is likely to further improve the effectiveness of SSM by better mask refining methods, \emph{e.g.}, improving the updating strategy for multi-scale masks and designing better refinement stopping criterion. The way to locate the targeted regions with a mask refinement module can also be applied to more unsupervised learning fields, opening avenues for future research.


\bibliographystyle{IEEEtran}
\bibliography{SSM.bib}
\vspace{-1cm}
\begin{IEEEbiography}[{\includegraphics[width=1in,height=1.25in,clip,keepaspectratio]{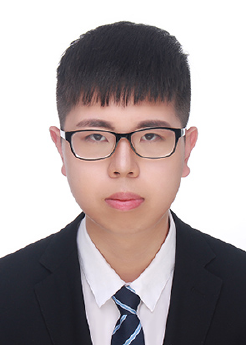}}]{Chaoqin Huang} (Student Member, IEEE) received his B.E. degree in computer science from Shanghai Jiao Tong University (SJTU), Shanghai, China, in 2019. He has been working towards a Ph.D. degree at the Cooperative Meidianet Innovation Center, Shanghai Jiao Tong University under the supervision of Prof. Ya Zhang, since 2019. He is also an intern at Shanghai AI Laboratory. His research interests include anomaly detection, computer vision, and machine learning.
\end{IEEEbiography}
\vspace{-1cm}
\begin{IEEEbiography}[{\includegraphics[width=1in,height=1.25in,clip,keepaspectratio]{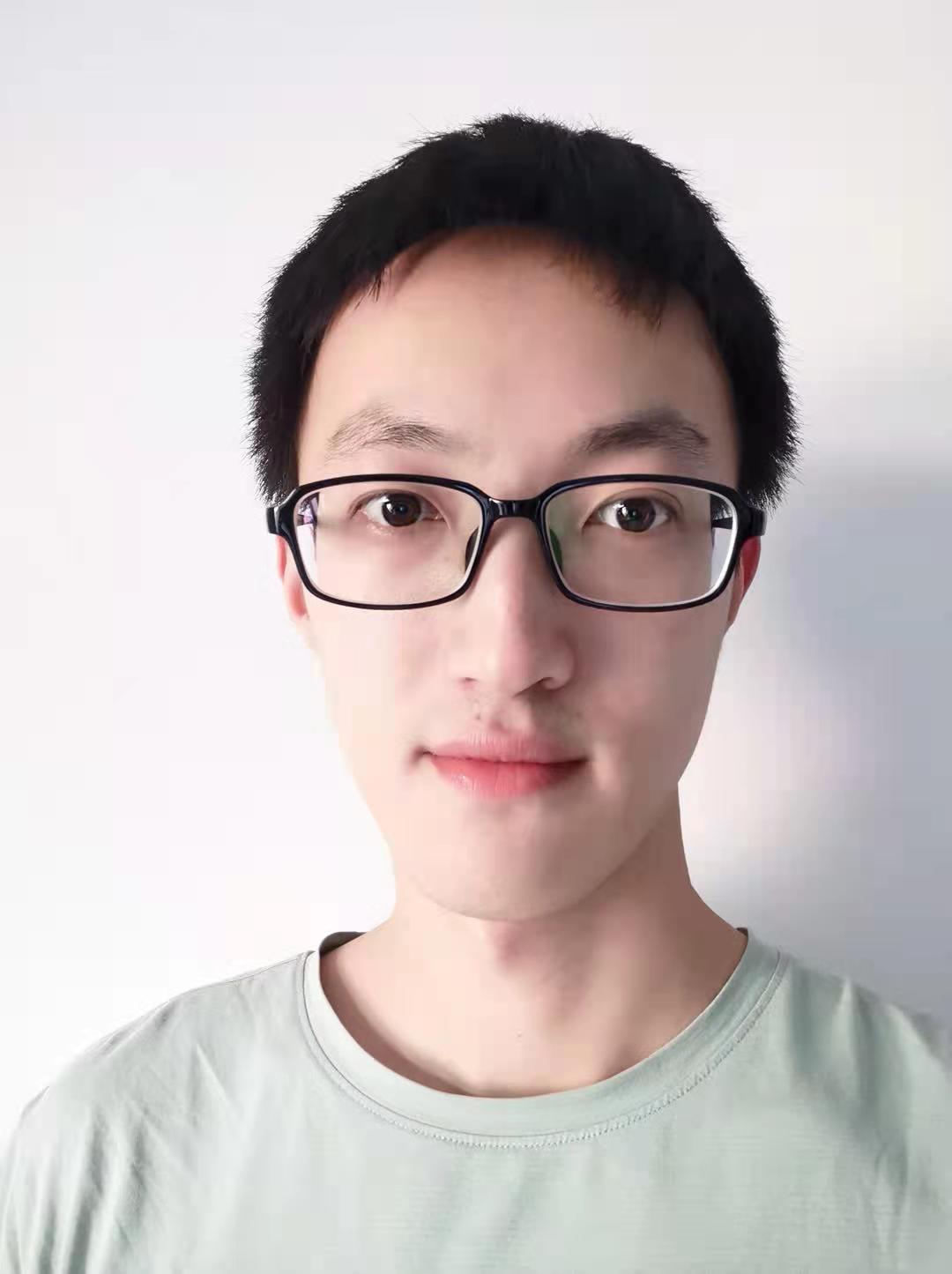}}]{Qinwei Xu} (Student Member, IEEE) received the B.E. degree in photoelectric information science and engineering from the University of Electronic Science and Technology of China, in 2018. He is currently pursuing the Ph.D. degree with the Cooperative Medianet Innovation Center, Shanghai Jiao Tong University, Shanghai, China, under the supervision of Prof. Ya Zhang. He is also an intern at Shanghai AI Laboratory. His research interests include deep learning and computer vision. 
\end{IEEEbiography}
\vspace{-1cm}
\begin{IEEEbiography}[{\includegraphics[width=1in,height=1.25in,clip,keepaspectratio]{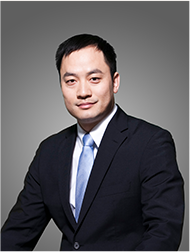}}]
{Yanfeng Wang} (Member, IEEE) received the B.S. degree from PLA Information Engineering University, and the master’s and Ph.D. degrees in business management from Shanghai Jiao Tong University. He is currently the Vice Director of Cooperative Medianet Innovation Center and also the Vice Dean of the School of Electrical and Information Engineering, Shanghai Jiao Tong University. His research interest is mainly on media big data, emerging commercial applications of information technology, and technology transfer.
\end{IEEEbiography}
\vspace{-1cm}
\begin{IEEEbiography}[{\includegraphics[width=1in,height=1.25in,clip,keepaspectratio]{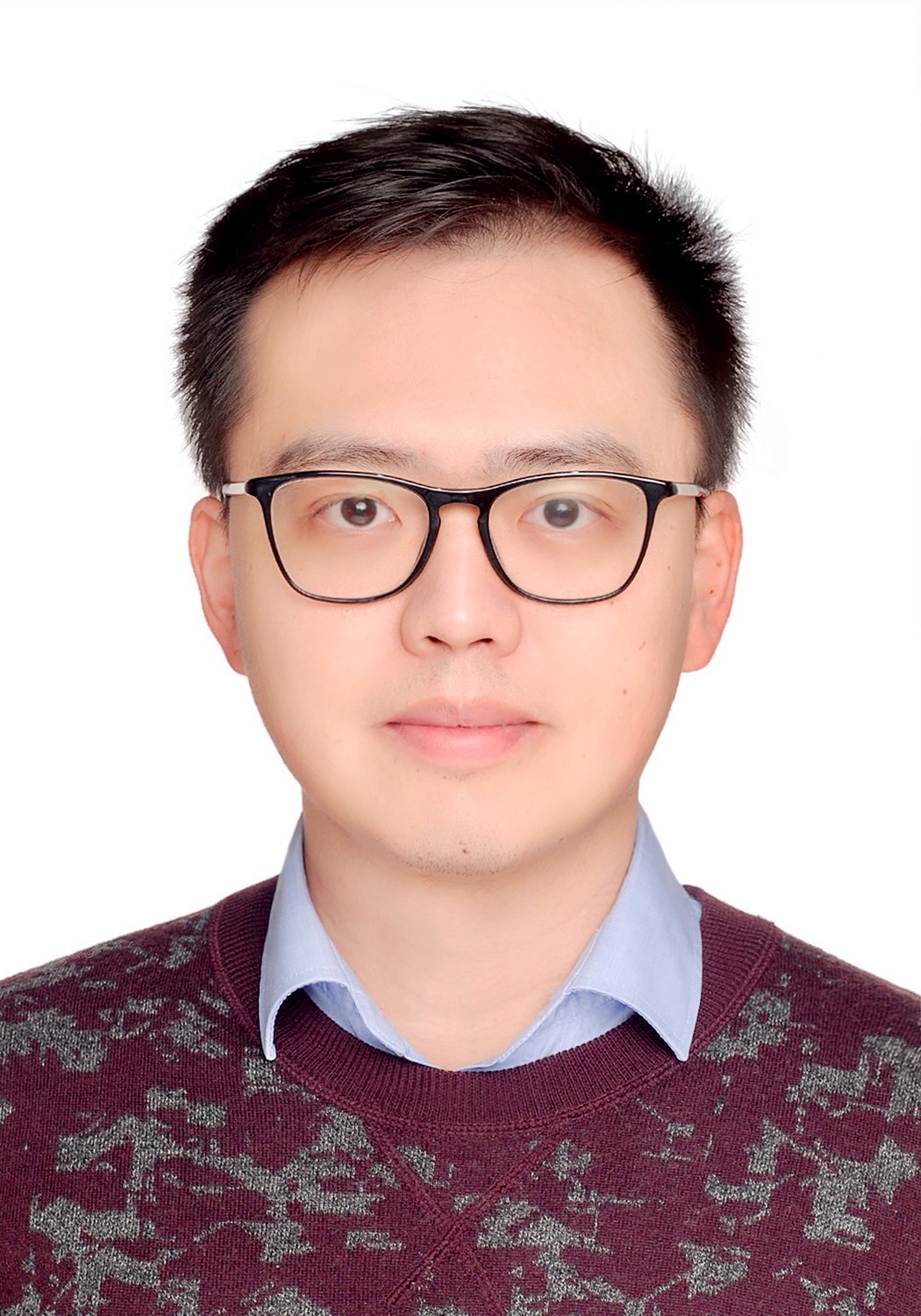}}]{Yu Wang} (Member, IEEE) received the B.E. degree from Huazhong University of Science and Technology, Wuhan, China, in 2009, and the M.Sc. degree in Communications and Signal processing and the Ph.D. degree in Signal Processing, both from Imperial College, London, U.K. in 2010 and 2015, respectively. He was a Senior Research Associate with Machine Intelligence Laboratory, Cambridge University Engineering Department, when he was a key member of the ALTA and IARPA MATERIAL projects. Since December 2020, he has been a Associate Professor with Cooperative Medianet Innovation Center, Shanghai Jiao Tong University. His current research interests include machine learning, audio signal processing,  speech recognition, spoken language processing and multi-modal signal processing. He is a Member of IEEE and ISCA.
\end{IEEEbiography}
\vspace{-1cm}
\begin{IEEEbiography}[{\includegraphics[width=1in,height=1.25in,clip,keepaspectratio]{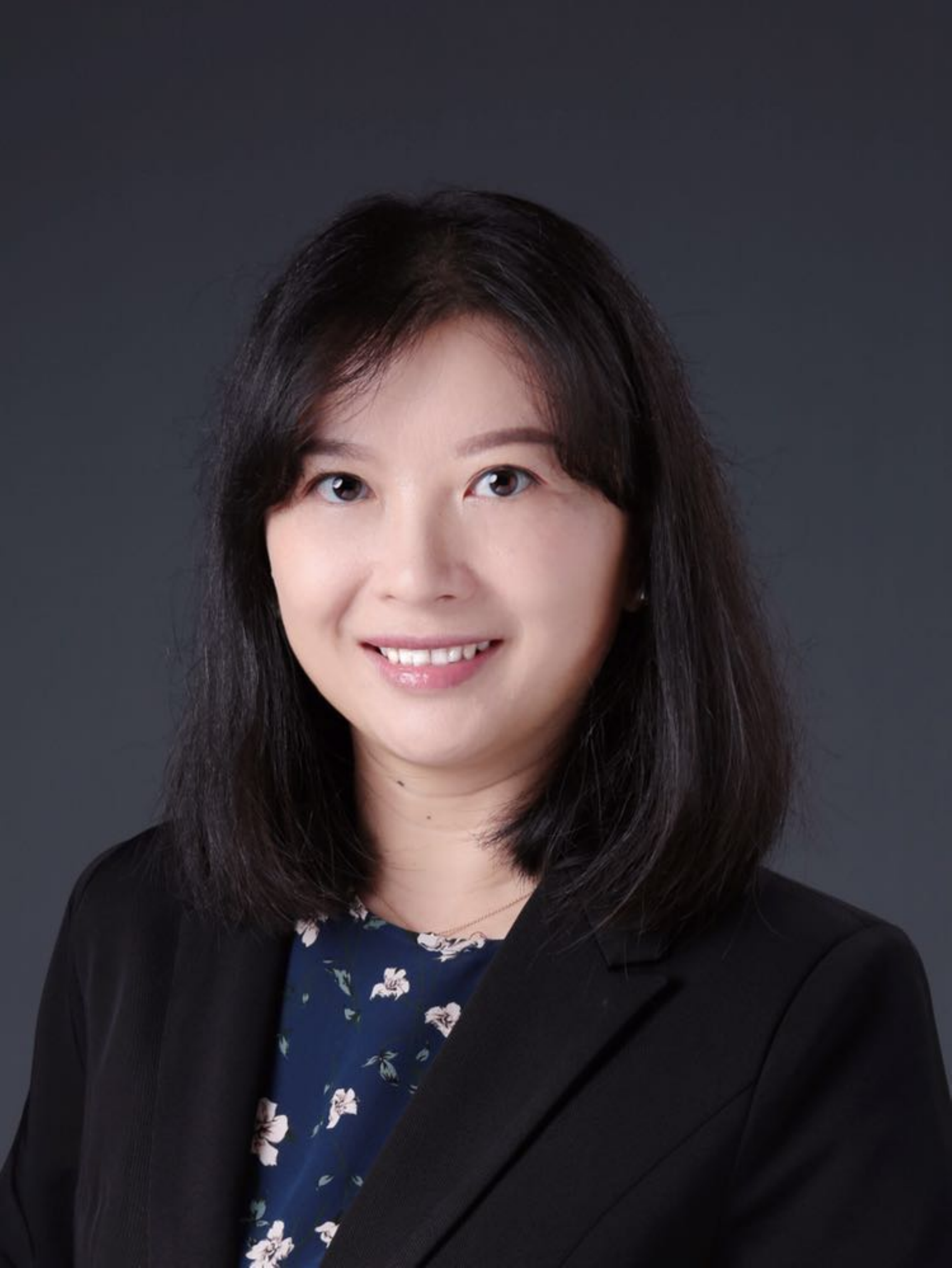}}]{Ya Zhang} (Member, IEEE)
is currently a professor at the Cooperative Medianet Innovation Center, Shanghai Jiao Tong University. Her research interest is mainly in machine learning with applications to multimedia and healthcare. Dr. Zhang has published more than 100 refereed papers in prestigious international conferences and journals. She has won several best paper awards of international journals and conferences, and directed one Outstanding Doctorate Dissertations awarded by Chinese Association for Artificial Intelligence.
\end{IEEEbiography}

\newpage
\onecolumn
\appendix
\renewcommand \arraystretch{1.0}
\begin{table*}[!h]
\centering
\caption{Complete results of ablation studies on different grid sizes of masks for SSM about \textbf{anomaly detection} on the MVTec AD dataset. Results are listed as AUC in \% and are marked individually for each class. An average score over all classes is also reported in the last row. The best-performing method is in bold.}
\label{tal:mvtec_com1}
\small
\setlength{\tabcolsep}{0.8pt}{
\begin{tabular}{C{1.7cm}C{1.4cm}C{1.4cm} C{1.4cm} C{1.4cm} C{1.4cm} C{1.4cm}C{1.4cm} C{1.6cm} C{1.6cm} C{1.9cm}}
\toprule
\multirow{2}{*}{Category} & \multicolumn{10}{c}{Grid Sizes of Masks}\\
& $\left[4\right]$ &
$\left[8\right]$ &
$\left[16\right]$ &
$\left[32\right]$ &
$\left[4,8\right]$ &
$\left[8,16\right]$ &
$\left[16,32\right]$ &
$\left[4,8,16\right]$ &
$\left[8,16,32\right]$ &
$\left[4,8,16,32\right]$\\
\cmidrule(lr){1-1} \cmidrule(lr){2-2} \cmidrule(lr){3-3} \cmidrule(lr){4-4} \cmidrule(lr){5-5} \cmidrule(lr){6-6} \cmidrule(lr){7-7} \cmidrule(lr){8-8} \cmidrule(lr){9-9} \cmidrule(lr){10-10} \cmidrule(lr){11-11}
Bottle & 98.2 & 98.7 & 99.5 & 99.5 & 99.7 & 99.8 & 100 & 99.9 & 99.9 & 100 \\
Cable & 59.1 & 60.8 & 67.3 & 76.6 & 74.3 & 74.8 & 77.5 & 77.3 & 76.3 & 77.2 \\
Capsule & 87.2 & 86.5 & 90.0 & 90.5 & 89.7 & 91 & 91.6 & 91.4 & 92.5 & 91.7 \\
Carpet & 67.7 & 67.9 & 76.8 & 72.7 & 72.9 & 74.8 & 73.9 & 76.3 & 73.3 & 73.4 \\
Grid & 94.2 & 100 & 100 & 99.9 & 100 & 100 & 100 & 100 & 100 & 100 \\
Hazelnut & 81.6 & 88.5 & 92.2 & 95.4 & 88.1 & 91.5 & 93.5 & 91.5 & 95.9 & 93.5 \\
Leather & 99.0 & 100 & 99.8 & 99.4 & 100 & 99.7 & 99.8 & 99.9 & 99.3 & 99.0 \\
Metal Nut & 90.2 & 74.2 & 90.3 & 88.8 & 76.7 & 86.9 & 89.6 & 88.7 & 91.7 & 90.2 \\
Pill & 94.4 & 92.1 & 89.7 & 83.9 & 94.2 & 89.6 & 82.8 & 89.1 & 82.1 & 85.6 \\
Screw & 80.5 & 86.5 & 83.8 & 82.1 & 87.5 & 85.3 & 81.0 & 85.0 & 82.3 & 83.5 \\
Tile & 97.7 & 96.1 & 94.9 & 89.6 & 95.9 & 93.3 & 91.0 & 94.4 & 89.6 & 96.6 \\
Toothbrush & 96.7 & 98.3 & 100 & 100 & 99.4 & 100 & 100 & 100 & 100 & 100 \\
Transistor & 74.0 & 82.5 & 86.0 & 93.5 & 84.9 & 90.3 & 95.5 & 91.0 & 93.5 & 92.7 \\
Wood & 98.0 & 98.2 & 93.8 & 90.6 & 98.7 & 94.5 & 94.5 & 95.9 & 94.9 & 95.2 \\
Zipper & 99.9 & 99.9 & 99.9 & 99.6 & 99.9 & 100 & 99.6 & 99.9 & 99.6 & 99.8 \\
\cmidrule(lr){1-1} \cmidrule(lr){2-2} \cmidrule(lr){3-3} \cmidrule(lr){4-4} \cmidrule(lr){5-5} \cmidrule(lr){6-6} \cmidrule(lr){7-7} \cmidrule(lr){8-8} \cmidrule(lr){9-9} \cmidrule(lr){10-10} \cmidrule(lr){11-11}
Mean & 87.5 & 89.3 & 90.9 & 90.8 & 90.8 & 91.4 & 91.3 & \textbf{92.0} & 91.4 & 91.9 \\
\bottomrule
\end{tabular}}
\end{table*}

\renewcommand \arraystretch{1.0}
\begin{table*}[!h]
\centering
\caption{Complete results of ablation studies on different grid sizes of masks for SSM about \textbf{anomaly localization} on the MVTec AD dataset. Results are listed as AUC in \% and are marked individually for each class. An average score over all classes is also reported in the last row. The best-performing method is in bold.}
\label{tal:mvtec_com2}
\small
\setlength{\tabcolsep}{0.8pt}{
\begin{tabular}{C{1.7cm}C{1.4cm}C{1.4cm} C{1.4cm} C{1.4cm} C{1.4cm} C{1.4cm}C{1.4cm} C{1.6cm} C{1.6cm} C{1.9cm}}
\toprule
\multirow{2}{*}{Category} & \multicolumn{10}{c}{Grid Sizes of Masks}\\
& $\left[4\right]$ &
$\left[8\right]$ &
$\left[16\right]$ &
$\left[32\right]$ &
$\left[4,8\right]$ &
$\left[8,16\right]$ &
$\left[16,32\right]$ &
$\left[4,8,16\right]$ &
$\left[8,16,32\right]$ &
$\left[4,8,16,32\right]$\\
\cmidrule(lr){1-1} \cmidrule(lr){2-2} \cmidrule(lr){3-3} \cmidrule(lr){4-4} \cmidrule(lr){5-5} \cmidrule(lr){6-6} \cmidrule(lr){7-7} \cmidrule(lr){8-8} \cmidrule(lr){9-9} \cmidrule(lr){10-10} \cmidrule(lr){11-11}
Bottle & 92.2 & 93.8 & 95.6 & 96.1 & 94.1 & 96.0 & 96.4 & 95.9 & 96.5 & 96.6 \\
Cable & 81.1 & 78.9 & 75.8 & 81.2 & 78.9 & 79.2 & 81.3 & 82.1 & 77.4 & 80.9 \\
Capsule & 93.4 & 95.1 & 98.0 & 98.4 & 97.3 & 98.1 & 98.7 & 98.4 & 98.5 & 98.5 \\
Carpet & 90.4 & 87.1 & 94.2 & 92.0 & 92.8 & 91.5 & 92.1 & 94.4 & 92.9 & 92.9 \\
Grid & 98.5 & 99.0 & 98.9 & 98.7 & 99.0 & 99.0 & 98.8 & 99.0 & 98.8 & 98.8 \\
Hazelnut & 96.3 & 96.9 & 97.7 & 98.2 & 96.9 & 97.8 & 98.1 & 97.4 & 98.0 & 98.1 \\
Leather & 99.5 & 99.7 & 99.6 & 99.6 & 99.6 & 99.6 & 99.6 & 99.6 & 99.6 & 99.6 \\
Metal Nut & 77.8 & 81.4 & 89.1 & 93.2 & 83.7 & 89.5 & 92.4 & 89.6 & 93.2 & 92.8 \\
Pill & 99.1 & 98.8 & 98.0 & 96.5 & 98.8 & 97.6 & 96.7 & 97.8 & 96.6 & 96.6 \\
Screw & 98.0 & 98.9 & 98.9 & 98.8 & 99.0 & 99.0 & 98.9 & 98.9 & 99.0 & 99.0 \\
Tile & 96.6 & 93.5 & 88.4 & 81.9 & 93.4 & 88.7 & 79.1 & 90.2 & 83.8 & 87.7 \\
Toothbrush & 98.5 & 98.8 & 98.9 & 98.9 & 98.8 & 98.8 & 98.9 & 98.9 & 98.9 & 98.9 \\
Transistor & 76.4 & 77.7 & 79.8 & 83.2 & 78.5 & 80.1 & 83.7 & 80.1 & 82.9 & 83.1 \\
Wood & 89.0 & 88.7 & 87.0 & 84.7 & 88.4 & 87.1 & 83.7 & 86.9 & 83.7 & 84.1 \\
Zipper & 99.0 & 99.0 & 99.0 & 98.9 & 99.0 & 99.0 & 98.9 & 99.0 & 98.9 & 98.9 \\
\cmidrule(lr){1-1} \cmidrule(lr){2-2} \cmidrule(lr){3-3} \cmidrule(lr){4-4} \cmidrule(lr){5-5} \cmidrule(lr){6-6} \cmidrule(lr){7-7} \cmidrule(lr){8-8} \cmidrule(lr){9-9} \cmidrule(lr){10-10} \cmidrule(lr){11-11}
Mean & 92.8 & 92.5 & 93.3 & 93.4 & 93.2 & 93.4 & 93.1 & \textbf{93.9} & 93.2 & 93.8 \\
\bottomrule
\end{tabular}}
\end{table*}

\begin{figure*}[!h]
    \centering
    \includegraphics[width=0.9\textwidth]{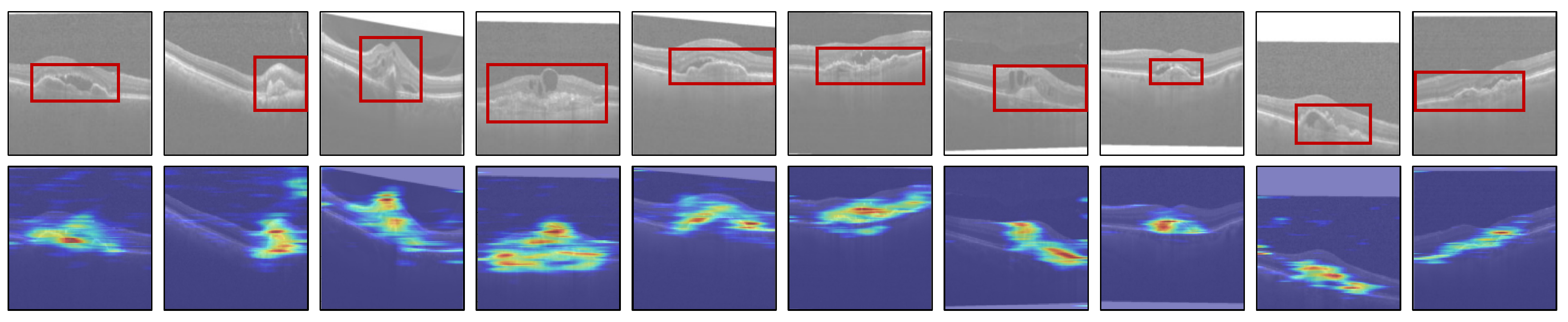}
    \caption{More qualitative results of anomaly localization with SSM on the Retinal-OCT dataset.}
    \label{fig:retinal_all}
\end{figure*}

\newpage
To better show the main contributions in this paper, we summarize the differences between the traditional image reconstruction-based method and the image inpainting-based method in Table~\ref{tal:sum}.
\begin{table}[h]
\normalsize
\centering
\setlength{\tabcolsep}{4.5pt}{
\begin{tabular}{L{2.5cm}|L{5.5cm}|L{7.5cm}}
\hline
& Without Masking & With Masking\\
\hline
Framework & Image Reconstruction & Image Restoration / Image Inpainting \\
\hline
Inputs & Entire Image & Image + Mask \\
\hline
Assumption & The model is enforced to learn regularities of normal data to minimize reconstruction errors; anomalies are difficult to be reconstructed and thus have large reconstruction errors. & The model is enforced to predict the missing information from its surrounding normal regions; the difference between the original masked region and its corresponding restoration result is significant for anomalous regions.\\
\hline
Challenge & Compressing the original image and then reconstructing it. & The dual tasks of \emph{masking} possible anomalies and \emph{restoring} the masked regions. \\
\hline
Pros/Cons & {\begin{itemize}[leftmargin=0.3cm] 
    \item No need for mask design and mask selection. 
    \item Learned representation may focus on low level details.
\end{itemize}} & 
{\begin{itemize}[leftmargin=0.3cm]
    \item Need careful design and selection of the mask. 
    \item If the mask lies in the anomalous regions correctly, the image inpainting-based method performs significantly better than the image reconstruction-based method.
\end{itemize}}
\\
\hline
\end{tabular}}
\caption{Summarization of the key contribution of the proposed method.}\label{tal:sum}
\end{table}

\ifCLASSOPTIONcaptionsoff
  \newpage
\fi
\end{document}